\documentclass[11pt]{article}

\usepackage[preprint]{acl}

\usepackage{times}
\usepackage{latexsym}
\usepackage[T1]{fontenc}
\usepackage[utf8]{inputenc}
\usepackage{microtype}
\usepackage{inconsolata}
\usepackage{graphicx}
\usepackage{enumitem}
\usepackage{xspace}
\usepackage{amsmath}
\usepackage{amssymb}
\usepackage{colortbl}
\usepackage{booktabs}
\usepackage{multirow}
\usepackage{graphicx}
\usepackage{xcolor}
\usepackage{geometry}
\usepackage{algorithm}
\usepackage{algorithmic}
\usepackage{subcaption}
\usepackage{tcolorbox}

\newcommand{\eagle}{\textsc{Eagle}\xspace}
\newcommand{\method}{\textbf{\textsc{Talon}}\xspace}
\definecolor{desc}{RGB}{99, 178, 238}
\definecolor{acc}{RGB}{118, 218, 145}
\definecolor{rej}{RGB}{248, 149, 136}
\definecolor{uclablue}{rgb}{0.15,0.45,0.68}

\definecolor{sharedColor}{gray}{0.55}
\definecolor{talonColor}{HTML}{1F77B4}
\definecolor{eagleColor}{HTML}{E15759}

\definecolor{sagegreen}{RGB}{235, 247, 240}
\newcommand{\green}{\cellcolor{sagegreen}}

\definecolor{tok_acc_bg}{HTML}{E0F7FA} 
\definecolor{tok_acc_fg}{HTML}{00838F} 
\definecolor{tok_rej_bg}{HTML}{FCE4EC} 
\definecolor{tok_rej_fg}{HTML}{AD1457} 
\newcommand{\tokacc}[1]{\colorbox{tok_acc_bg}{\textcolor{tok_acc_fg}{\ttfamily #1}}}
\newcommand{\tokrej}[1]{\colorbox{tok_rej_bg}{\textcolor{tok_rej_fg}{\ttfamily #1}}}

\newcommand{\toktxt}[1]{\textcolor{darkgray}{\ttfamily #1}}

\title{
    \raisebox{-0.2\height}{\includegraphics[width=0.1\textwidth]{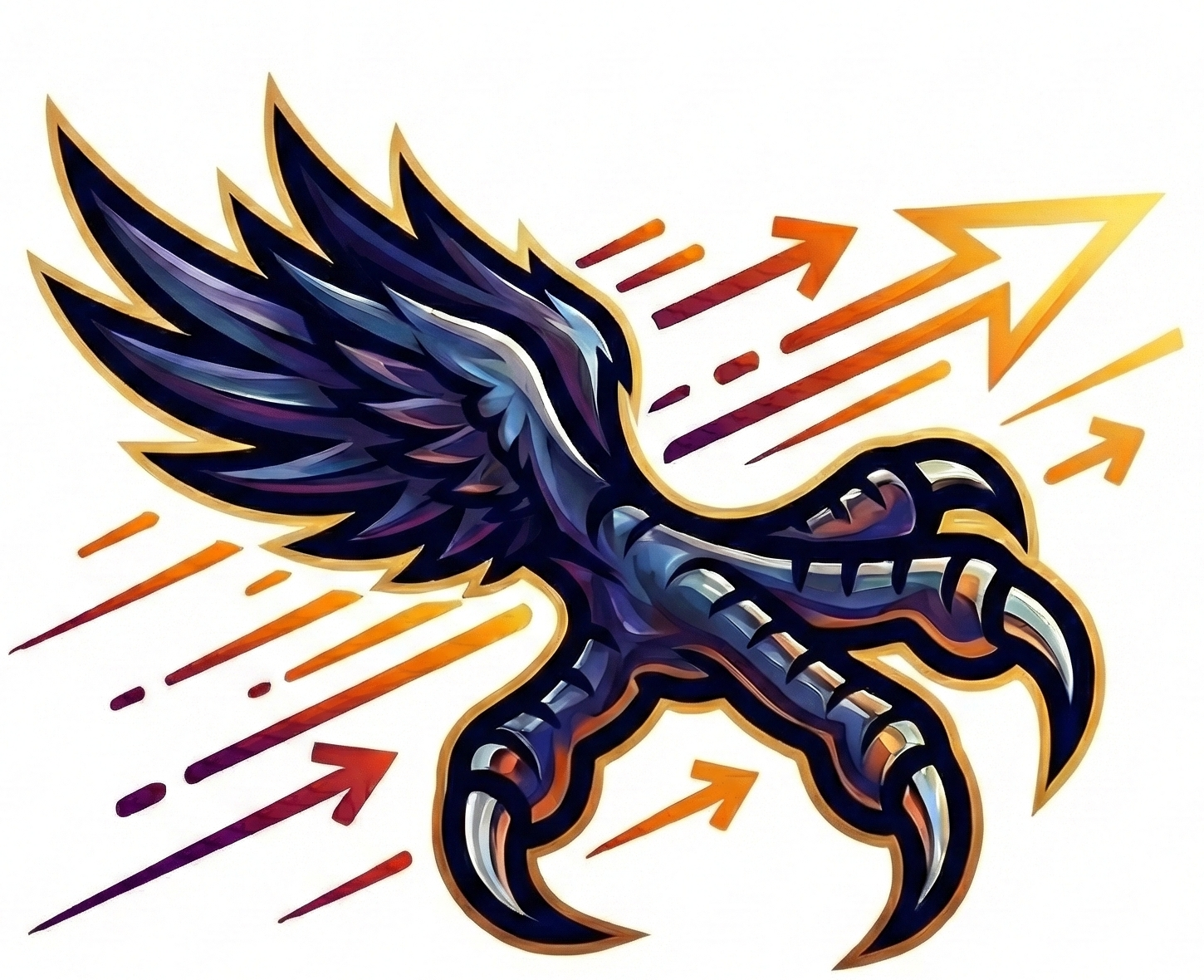}} 
    \method: Confidence-Aware Speculative Decoding with Adaptive Token Trees    
}

\author{
   Tianyu Liu$^{1,2}$\thanks{Equal Contribution.} \quad Qitan Lv$^{1,2*}$ \quad Yuhao Shen$^{3}$ \quad Xiao Sun$^{2}$\thanks{The Corresponding Author.} \quad Xiaoyan Sun$^{1}$ \quad 
  \\
  ${}^1$University of Science and Technology of China \\
  ${}^2$Shanghai AI Laboratory \quad
  ${}^3$Zhejiang University\\
    \texttt{\{tianyu\_liu, qitanlv\}@mail.ustc.edu.cn} \quad \texttt{riven@zju.edu.cn}\\
    \texttt{sunxiao@pjlab.org.cn}
    \quad \texttt{sunxiaoyan@ustc.edu.cn}
}

\begin{document}
\maketitle

\begin{abstract}
    Speculative decoding (SD) has become a standard technique for accelerating LLM inference without sacrificing output quality.
    Recent advances in speculative decoding have shifted from sequential chain-based drafting to tree-structured generation, where the draft model constructs a tree of candidate tokens to explore multiple possible drafts in parallel.
    However, existing tree-based SD methods typically build a \textbf{fixed-width, fixed-depth} draft tree, which fails to adapt to the varying difficulty of tokens and contexts.
    As a result, the draft model cannot dynamically adjust the tree structure to early stop on difficult tokens and extend generation for simple ones.
    To address these challenges, we introduce \method, a \textbf{\textit{training-free, budget-driven}} adaptive tree expansion framework that can be plugged into existing tree-based methods.
    Unlike static methods, \method constructs the draft tree iteratively until a fixed token budget is met, using a hybrid expansion strategy that adaptively allocates the node budget to each layer of the draft tree.
    This framework naturally shapes the draft tree into a \textbf{``deep-and-narrow''} form for deterministic contexts and a \textbf{``shallow-and-wide''} form for uncertain branches, effectively optimizing the trade-off between exploration width and generation depth under a given budget.
    Extensive experiments across 5 models and 6 datasets demonstrate that \method consistently outperforms state-of-the-art \eagle-3, achieving up to 5.16$\times$ end-to-end speedup over auto-regressive decoding.
\end{abstract}







\section{Introduction}
\label{sec:intro}

\begin{figure}[t]
    \centering
    \includegraphics[width=\columnwidth]{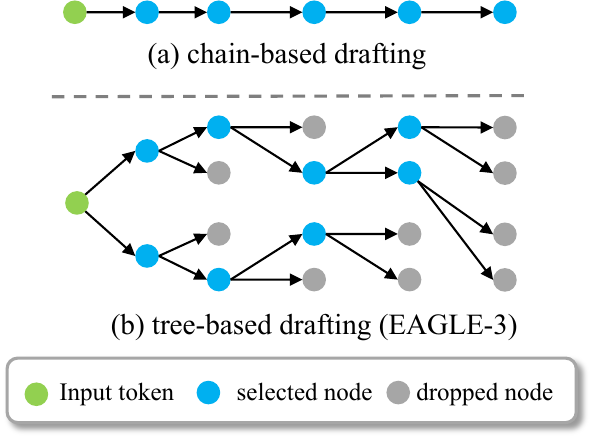}
    \caption{Illustration of chain-based drafting and tree-based drafting (\eagle-3~\citep{eagle3}) with $K=2$. At each step, \eagle calls draft model forward on the selected $K$ parent nodes of last step, selects top-$K$ child nodes for each parent, and filters $K\times K$ child nodes. Then \eagle employs an additional top-$K$ operation to choose $K$ nodes as parents for next step.}
    \label{fig:intro-1}
\end{figure}

While Large Language Models (LLMs) have achieved remarkable success in various benchmarks \citep{gpt-5, gemini-3, opus-4-5, qwen-3, deepseek-r1, glm}, their deployment is severely constrained by their auto-regressive token-by-token generation. 
This sequential dependency prevents models from predicting multiple tokens \citep{mtp} in a single step, causing inference latency to scale linearly with output length \citep{palm-scaling} and making real-time interaction computationally expensive.


To alleviate this limitation, \textit{Speculative Decoding} (SD) \citep{sps1, sps2} has emerged as a promising paradigm to break the strict sequential dependency \citep{pearl}. 
SD decouples each decoding step into two sub-procedures: efficient drafting and parallel verification. A lightweight draft model first proposes a short candidate sequence, and the target LLM then validates all proposed tokens in parallel, accepting multiple tokens with a single target model forward.

However, early speculative decoding methods typically employ a \textit{chain-based} drafting strategy, which is inherently vulnerable. 
A single rejection at the beginning of the draft sequence invalidates all subsequent tokens, leading to a substantial waste of computational resources \citep{specinfer}. 
To further improve the overall acceptance rate, recent works have shifted from chain-based to \textit{tree-based drafting} \citep{specinfer, medusa, eagle, glide, eagle2}. 
Instead of predicting a single sequence, these approaches construct a token tree covering multiple plausible continuation paths. 
By leveraging \textit{Tree Attention} \citep{flash-tree-attn}, the target LLM can verify multiple draft token sequences in parallel within a single forward pass, significantly increasing the draft acceptance rate and overall acceleration.

\begin{figure}[t]
    \centering
    \includegraphics[width=\columnwidth]{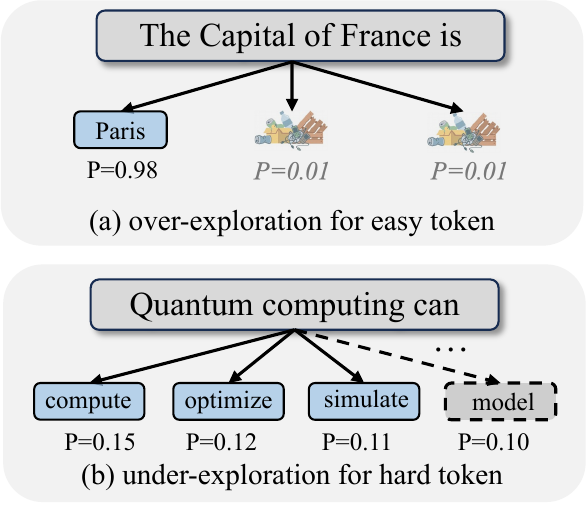}
    \caption{Limitations of tree-based drafting methods. (a) when the model is already confident to its prediction, the draft tree still grows $k$ child nodes. (b) when the model is very confused and highly uncertain, the top-$K$ draft tokens are still not sufficient.}
    \label{fig:intro-2}
\end{figure}

State-of-the-art tree-based methods, such as \eagle \citep{eagle}, construct the draft tree via a rigid layer-wise expansion mechanism. 
As illustrated in Figure~\ref{fig:intro-1}, these approaches iteratively expand a fixed number of child nodes for every parent and select the top-$K$ candidates to proceed to the next layer. 
However, this strategy enforces a \textit{static tree structure} with pre-determined dimensions. 
Such rigid allocation fails to adapt to the model's dynamic confidence (output probabilities of its tokens) \footnote{``Dynamic'' in \eagle means that the draft model dynamically select top-$K$ nodes as next-layer parents, but it cannot adjust $k$ to adapt for contexts. We provide an example in Figure~\ref{fig:eagle2_examples} in Appendix to further illustrate the difference.}: it still expands $k$ child nodes even if the draft model is already confident (shown in Figure~\ref{fig:intro-2}(a)), while it cannot allocate more node budget when the model is confused and highly uncertain (shown in Figure~\ref{fig:intro-2}(b)). Moreover, it prematurely truncates high-confidence paths that could extend deeper, while squandering the computational budget on expanding low-probability branches in uncertain contexts. 

To tackle these inefficiencies, we introduce \method, a novel \textit{training-free, budget-driven} adaptive tree expansion framework to op\textbf{T}imize dr\textbf{A}ft tree for faster specu\textbf{L}ative dec\textbf{O}di\textbf{N}g.
Departing from the rigid fixed width and depth generation, \method employs a dynamic growth algorithm that incrementally expands the draft tree until the number of nodes reaches a given budget $N$.
Specifically, we design a hybrid expansion strategy to optimize the tree topology: 
at the first layer, we propose a fixed width initialization that utilizes a top-$K$ operation to alleviate the early rejection phenomenon. At subsequent layers, we propose a confidence-gated expansion strategy to adaptively allocate the node budget.
This allows the draft structure to evolve to \textbf{deep-and-narrow} for deterministic contexts to maximize draft length, or \textbf{shallow-and-wide} for uncertain ones to enhance hit rate. 

To summarize, our contributions are:
\begin{enumerate}
    \item[(i)] We identify a key limitation in existing tree-based speculative decoding methods: the strict, layer-wise static tree expansion fails to adapt to the model's varying confidence, leading to inefficient utilization of the computational budget.
    \item[(ii)] We propose \method, a budget-driven speculative decoding framework. By employing a dynamic tree growth algorithm with robust tree initialization and confidence-gated expansion, \method constructs adaptive draft trees that dynamically adapts (deep-and-narrow or shallow-and-wide) based on context uncertainty.
    \item[(iii)] Extensive experiments verify the effectiveness of \method. It significantly outperforms state-of-the-art method \eagle-3 in terms of draft efficiency and wall-clock speedup, particularly in scenarios with fluctuating generation difficulty.
\end{enumerate}

\section{Related Work}\label{related_work}
In this section, we review speculative decoding from the perspective of \emph{drafting structures}: chain-based speculative decoding and tree-based speculative decoding.
\textbf{We position \method in the context of extensive related works in Appendix~\ref{app:related_work}.}

\paragraph{Chain-based speculative decoding.}
Speculative decoding \citep{sps1, sps2} introduces a lossless \emph{draft-and-verify} \citep{draft_and_verify} principle: a cheap drafter proposes a short continuation and a target model verifies it in parallel, accepting the longest matched prefix.
This paradigm suffers from a clear efficiency bottleneck---\emph{early rejection} \citep{block_verification}---since a mismatch at an early position invalidates all downstream drafted tokens.
Follow-up work improves chain-based SD by (i) producing better proposals with minimal overhead (e.g., lightweight head-based drafting that reuses target representations\citep{medusa, glide, eagle}), and (ii) reducing wasted draft computation via adaptive lookahead \citep{specdec++} or pipelined scheduling \citep{pearl}.
\textbf{Nevertheless, the speedup is fundamentally sensitive to the hardest tokens, because one rejection discards the entire suffix.}

\paragraph{Tree-based speculative decoding.}
Tree-based SD generalizes the draft from a single chain to a \emph{token tree}, allowing the verifier to choose among multiple candidate branches within one forward and thus mitigating early rejection.
SpecInfer \citep{specinfer} is an early representative that organizes candidates as a token tree and verifies them in parallel with tree attention.
More recent work strengthens the tree-based pipeline from two angles: (i) improving tree proposals via well-calibrated drafters \citep{adaeagle, c2t, falcon} and context-aware dynamic draft trees \citep{dyspec}, and (ii) optimizing the draft-tree structure under a node budget, including dynamic-programming-based and search-based construction (e.g., Sequoia \citep{sequoia} and OPT-Tree \citep{opt-tree}).
However, existing methods typically rely on \textbf{rigid or heuristic expansion patterns}, failing to \textbf{explicitly allocate a token budget based on real-time confidence.}
This gap motivates \method, a training-free, budget-driven framework that constructs adaptive token trees (deep-and-narrow vs.\ shallow-and-wide).

\section{Background}
\label{sec:background}

We first formulate \textit{chain-based} drafting and \textit{tree-based} drafting for better understanding.

\subsection{Chain-based Drafting}
In standard speculative decoding, a smaller draft model $\mathcal{M}_d$ generates a single sequence of $\gamma$ tokens (a chain) as a speculation for the target model $\mathcal{M}_t$. Given the prefix $x_{\le t}$, the drafting process generates a sequence $x_{t+1}, \dots, x_{t+\gamma}$ auto-regressively:
\begin{equation}
    x_{t+i} \sim P_{\mathcal{M}_d}(\cdot | x_{<t+i}), \quad 1 \le i \le \gamma
\end{equation}
The target model $\mathcal{M}_t$ then verifies this sequence in parallel. The key limitation of this approach is the \textit{sequential dependency} during verification: if the token at position $i$ is rejected, all subsequent tokens $x_{>i}$ are discarded regardless of their correctness, resulting in wasted computation.

\subsection{Tree-based Drafting}

\begin{figure}[t]
    \centering
    \includegraphics[width=\linewidth]{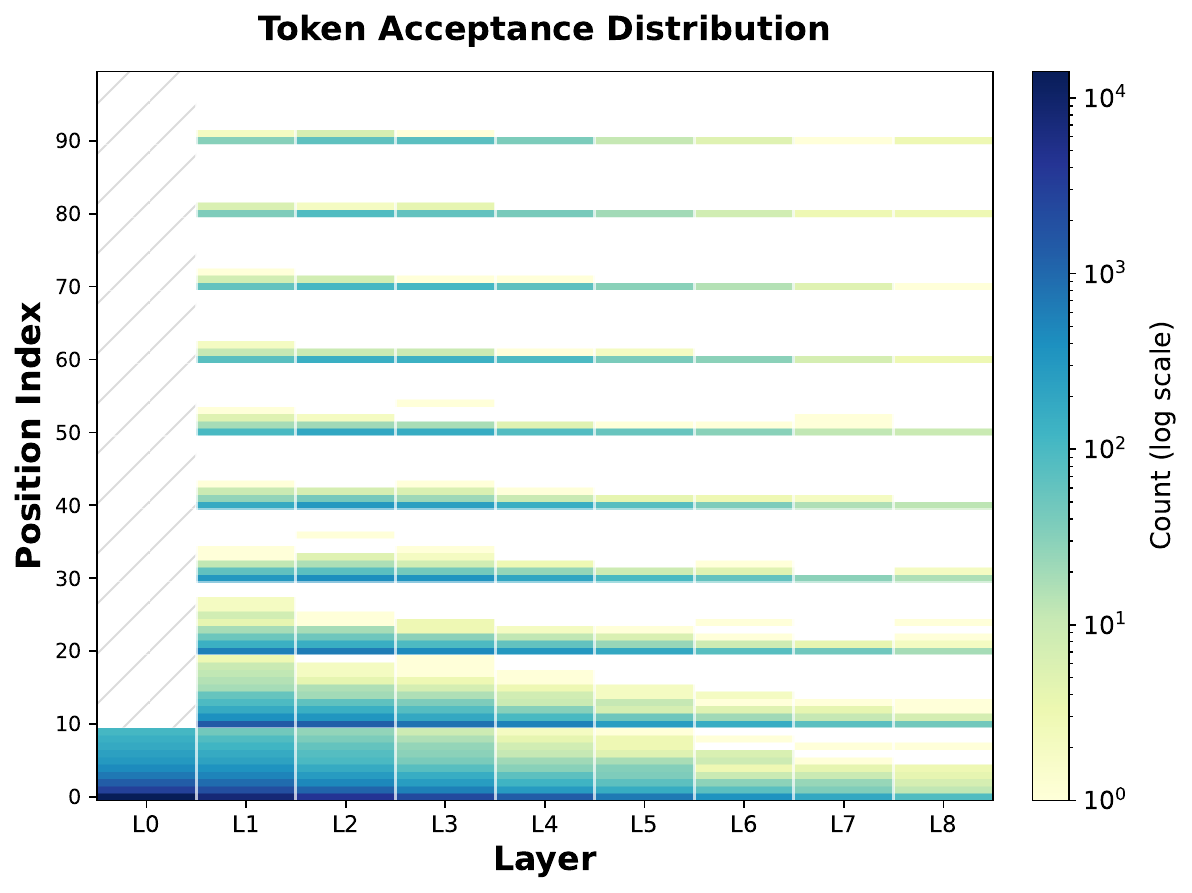}
    \caption{Visualization of token acceptance frequency within a static draft tree ($K=10, D=8$). The heatmap reveals an \textbf{"Acceptance Funnel"} effect: 
    while the acceptance frequency of the first layer is relatively uniform, the acceptance in subsequent layers ($d \ge 1$) shows a funnel trend that the accepted tokens concentrate more on high confidence regions (e.g., top-1 and top-2), rendering the wide static expansion computationally wasteful. Note that the first layer only has $K$ nodes, while its subsequent layers have $K\times K$ nodes.}
    \label{fig:funnel}
\end{figure}

To mitigate the limitations of chain-based drafting, recent works (e.g., \eagle) verify a token tree $\mathcal{T}$ to cover diverse paths. 
As formalized in Algorithm~\ref{alg:eagle} and Figure~\ref{fig:intro-1}, these methods employ a \textit{static} construction strategy: at each depth $d$, the model takes a parallel forward on all parent nodes and outputs their next-token distributions. Then, it selects top-$K$ entries from the output distribution of each parent node, acquiring $K\times K$ leaf nodes. After that, the model measures the path score of each layer node $v$:

\begin{equation}
   p(v)=\prod_{j \in \text{Path}(x_t, v)}  P_{\mathcal{M}_d} (j\ |\ x_{<j})
\end{equation}
where $\text{Path}(x_t, v)$ represents the path from the root node $x_t$ to the leaf node $v$, $x_{<j}$ denotes all prefix tokens of $j$. Then it uses another top-$K$ operation based on the path score to select $K$ leaf nodes as parent nodes of next layer. Finally, the generation ends at depth $D$, and the tree $\mathcal{T}$ will be pruned to meet a global budget $N$.
While structured, this rigid ``expand-then-shrink'' mechanism often generates redundant nodes that are discarded during intermediate shrinking or final pruning, leading to suboptimal resource allocation.

\section{Motivated Experiments}
\label{sec:motivation}

\begin{figure}[t]
    \centering
    \includegraphics[width=\linewidth]{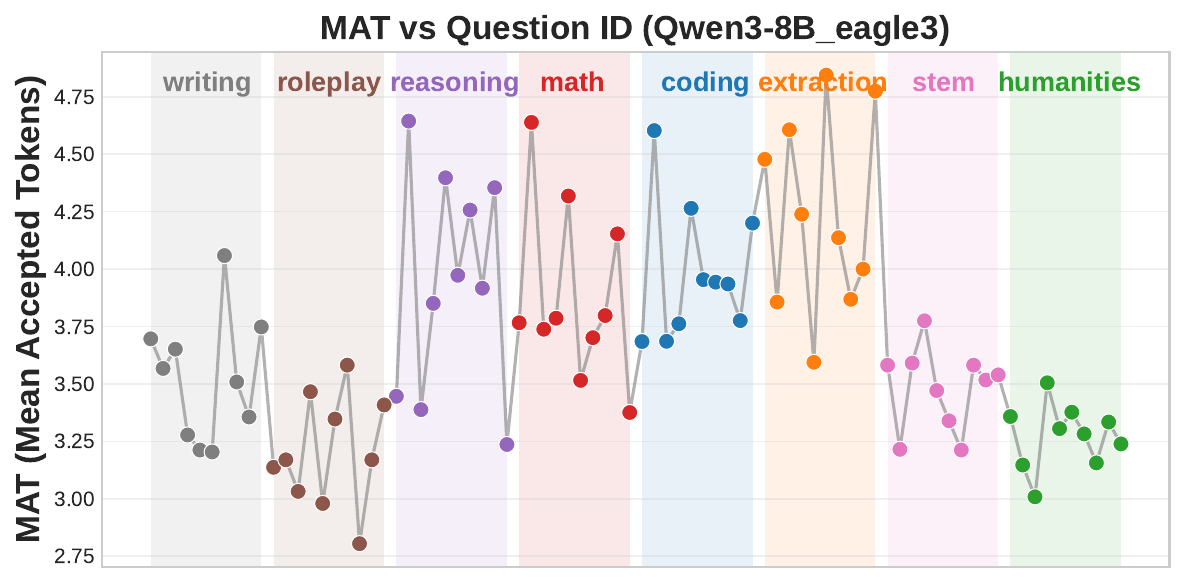}
    \caption{Real-World Mean Accepted Tokens (MAT) distribution across different queries in \eagle. The results exhibit high volatility: even within the same task category (e.g., Math or Coding), the optimal generation length fluctuates significantly.}
    \label{fig:dynamic_len}
\end{figure}

To empirically investigate the limitations of static tree-based speculative decoding, we conduct a pilot analysis using \eagle~\cite{eagle} as a representative baseline. We employ a fixed tree topology with width $K=10$ and depth $D=8$ (official settings in \eagle paper) with Qwen3-8B on MT-Bench \citep{mt-bench}. We also conduct additional motivated experiments in Appendix~\ref{app:motivated_exp} with Llama-series models and various datasets to demonstrate the generality of our motivation.
\subsection{The "Acceptance Funnel" Phenomenon}
\label{ssec:funnel}

We first investigate the acceptance frequency of each position in the draft tree. Detailed model configurations are provided in Table~\ref{tab:draft_models}. Figure~\ref{fig:funnel} visualizes the acceptance frequency of tokens at each position within the static tree structure. Two key observations emerge regarding the \textit{distribution of effective speculation}:

\noindent\textbf{Diminishing Returns of Width in Deep Layers.}
As illustrated in Figure~\ref{fig:funnel}, the acceptance distribution exhibits a funnel-like pattern. In the first initial layer ($d=0$), the acceptance probability is relatively uniform across the top-$K$ candidates. This suggests that due to the initial stochasticity and minor distributional divergence between the draft model $\mathcal{M}_d$ and the target model $\mathcal{M}_t$, a wider search breadth is necessary to capture the valid continuation. 
However, as the tree deepens ($d \ge 1$), the acceptance mass concentrates sharply on the high confidence regions (e.g. top-1 and top-2 candidates).
We attribute this to the \textit{cumulative error amplification} in auto-regressive drafting: for deep nodes, the draft model is either (1) confidently aligned with the target model (correct prediction), or (2) hallucinating a divergent path where even the top-$K$ candidates fail to recover the target distribution. Consequently, maintaining a static width $K=10$ at deeper layers yields negligible marginal utility. When the draft model is aligned, a much smaller $K$ can find the correct draft tokens. When the draft model is highly uncertain, there needs a larger $K$ to ensure the coverage and stops the generation to avoid wasteful draft model forward. 
\subsection{Variance in Real-World Accepted Length}
\label{ssec:dynamic_len}

While Section~\ref{ssec:funnel} reveals the structural redundancy within fixed width, we further investigate the weakness of fixed depth. We plot the distribution of Mean Accepted Tokens (MAT) for various queries in Figure~\ref{fig:dynamic_len}. (Same settings with Figure~\ref{fig:funnel})

\noindent\textbf{The Static Depth Dilemma.}
As shown in Figure~\ref{fig:dynamic_len}, the accepted tokens fluctuate drastically even within the same task category (e.g., Math or Coding), exposing the limitation of a fixed-depth drafting policy. In low-entropy contexts where the draft model is well-aligned with the target, the potential acceptance length often exceeds the pre-defined depth $D$. However, a static fixed depth $D$ prevents $\mathcal{M}_d$ from generating more draft tokens for maximum speedup. Conversely, in high-entropy scenarios where prediction becomes ambiguous, the draft model tends to hallucinate deep branches that are destined for rejection. A fixed depth $D$ forces the allocation of computational resources to these invalid tokens, incurring latency overhead with zero marginal utility.

\section{Method}
\label{sec:method}

We introduce \method, a budget-driven framework that optimizes draft tree construction by dynamically allocating resources based on real-time model confidence.

\begin{figure*}[t]
    \centering
    \includegraphics[width=\textwidth]{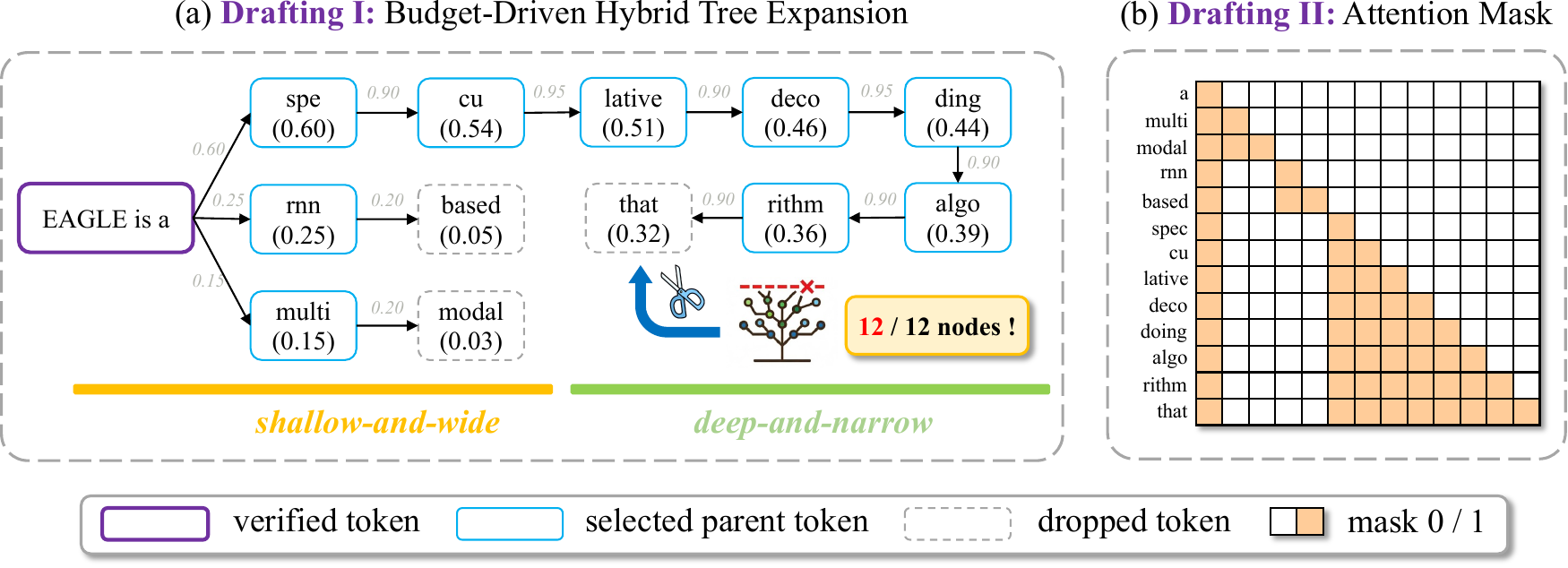}
    \caption{
(a) \method’s Budget-Driven Tree. \method dynamically allocates the token budget based on confidence. It uses Top-$K$ at the root for robustness and confidence gating at deeper layers, resulting in adaptive topologies—deep chains for high-confidence tokens (e.g., the folded “algo-rithm” sequence) and wide branches for uncertain ones. The expansion stops when the global budget is met (indicated by the icon).
(b) Tree Attention Mask. The structural mask used by the target model to verify the adaptive tree in parallel. The verification process follows the standard token-tree verification protocol (see Appendix~\ref{app:verification} for details).}
    \label{fig:talon-main}
\end{figure*}

\subsection{From Static Grids to Dynamic Budgets}
\label{sec:overview}

To overcome the inefficiencies of the \emph{static tree structures} analyzed in Section~\ref{sec:motivation}--specifically the misalignment between fixed topology and varying token difficulty--\method shifts the constraint from \emph{shape} to \emph{capacity}. We define a global \textbf{Token Budget} ($N$), representing the maximum number of nodes allowed in the draft tree $\mathcal{T}$. The objective is to iteratively ``invest'' this budget to grow a draft tree that dynamically adapts--\emph{deep-and-narrow} for deterministic contexts and \emph{shallow-and-wide} for uncertain ones--thereby maximizing effective speculation length under a fixed computational cost.

\subsection{Hybrid Tree Expansion Strategy}
\label{sec:hybrid-expansion}

\method constructs the tree layer-by-layer using a hybrid strategy: a robust initialization at the root (Layer 0) followed by confidence-gated expansion for subsequent layers (Layer $d \ge 1$).

\subsubsection{Confidence-Gated Expansion}
\label{sec:confidence-expansion}

For depths $d \ge 1$, we employ a global filtering mechanism to gate candidates based on their relative confidence. Let $\mathcal{P}_d$ be the set of parent nodes. We define the \textbf{Candidate Pool} $\mathcal{S}_d$ as the union of all child extensions:

\begin{equation}
    \mathcal{S}_d = \bigcup_{v \in \mathcal{P}_d} \{ (v, w) \mid w \in \mathcal{V} \}.
\end{equation}
Each candidate $u=(v, w) \in \mathcal{S}_d$ is assigned a cumulative path probability $p(u) = p(v) \cdot P_{\mathcal{M}_d}(w \mid x_{\le v})$.

Motivated by \citep{minp}, we first identify the \textbf{anchor confidence} $m_d = \max_{u \in \mathcal{S}_d} p(u)$. We then retain candidates whose confidence falls within a dynamic margin $\mu$ of the anchor:
\begin{equation}
    \mathcal{P}_{d+1} = \{ u \in \mathcal{S}_d \mid p(u) \ge \mu \cdot m_d \},
    \label{eq:gating}
\end{equation}
where $\mu \in (0, 1]$ is a hyperparameter. To strictly respect the budget, if $|\mathcal{P}_{d+1}|$ exceeds the remaining budget $N - |\mathcal{T}|$, we retain only the top candidates with the highest path scores.

\paragraph{Intuition.}
This mechanism naturally aligns topology with entropy. In deterministic contexts (e.g., ``The capital of France is \emph{Paris}''), the anchor $m_d \approx 1.0$ imposes a strict threshold, automatically pruning branches to form a \emph{deep-and-narrow} chain. In high-entropy contexts (e.g., ``Quantum computing can...''), a lower $m_d$ relaxes the threshold, admitting diverse candidates to form a \emph{shallow-and-wide} layer that prioritizes coverage.

\subsubsection{Robust Tree Initialization}
\label{sec:initialization}

While confidence gating effectively allocates budget at depth, applying it at the root ($d=0$) compromises robustness due to draft model \textbf{\textit{over-confidence}}.
Small draft models often exhibit imperfect calibration, assigning near-certain probability (e.g., $>0.99$) to incorrect tokens in simple contexts. Relative gating would prematurely collapse the tree into a single wrong branch--an \emph{over-confidence trap}. Unlike deeper errors, a root failure causes catastrophic rejection of the entire draft sequence.
To mitigate calibration errors, we employ a \textbf{Standard Top-$K$ Initialization} for the first layer. Regardless of the confidence distribution, we explicitly expand the top-$K$ tokens:
\begin{equation}
    \mathcal{P}_{1} = \{(x_t, v) \mid  v\in \text{Top-}K\left( P_{\mathcal{M}_d}(\cdot | x_{t}) \right)\}.
\end{equation}
This ensures the speculative tree covers diverse plausible directions at the most critical juncture before switching to budget-efficient gating.

\subsection{Evaluating \method with Draft Efficiency}
\label{sec:theoretical-analysis}

To rigorously quantify the cost-effectiveness of \method, we analyze the wall-time speedup by decomposing it into quality and cost factors. 

Let $N_p$ denote the total number of forward passes executed by the target model (verification steps), $N_q$ be the total forward passes executed by the draft model to generate a sequence and $L$ be the output length. We introduce a new metric, \textbf{Draft Efficiency} ($\delta$), defined as the ratio of these two quantities:
\begin{equation}
    \delta = \frac{N_q}{N_p}.
\end{equation}
Intuitively, $\delta$ represents the \textbf{speculation cost}--the average number of draft steps the system ``invests'' to secure a single verification opportunity. By deriving the wall-time speedup $R$ as a function of the Mean Accepted Tokens ($\tau= L/N_p$) and the relative model cost ($c$), we obtain the following speedup formulation ( see derivation in Appendix~\ref{app:derivation}):
\begin{equation}
    R = \frac{\tau}{1 + c \cdot \delta}.
    \label{eq:speedup_tradeoff}
\end{equation}
With Equation~\ref{eq:speedup_tradeoff}, we can theoretically explain \textbf{why \method achieves superior performance}. Approaches like \eagle enforce a fixed depth $D$, effectively locking the draft cost to a constant $\delta = D+1$ regardless of the context complexity. This rigidity leads to dual inefficiencies.

In \textbf{high-entropy contexts} (hard cases), the acceptance reward $\tau$ naturally drops. \eagle continues to pay the high fixed cost $\delta$, causing the speedup $R$ to degrade significantly. \method addresses this by halting expansion early; the resulting reduction in ``investment'' ($\delta \downarrow$) compensates for the lower reward, thereby mitigating the performance drop. Conversely, in \textbf{low-entropy contexts} (easy cases), static methods artificially cap the reward $\tau$ at depth $D$. \method allows the draft tree to extend far beyond this limit, significantly increasing $\tau$. While growing deeper also increases the draft cost $\delta$, this investment yields a positive net gain. Since the relative cost ratio is typically small ($c \ll 1$), $(1 + c \delta)$ grows much slower than $\tau$, ensuring that the overall wall-time speedup $R$ continues to improve as the tree deepens.
\section{Experiments}

\begin{table*}[htbp]
\centering
\renewcommand{\arraystretch}{1.1}
\caption{\textbf{Main Results on Six Benchmarks (Temperature=0).} Comparison between \eagle-3 and our proposed \textbf{\method} across various models. We report Mean Acceptance Tokens (MAT) and Wall-time Speedup relative to standard decoding. \textbf{Bold} numbers denote the best speedup performance.}
\label{tab:main_results}

\resizebox{\textwidth}{!}{%
\setlength{\tabcolsep}{4pt}
\begin{tabular}{llcccccccccccc}
\toprule
\multirow{2}{*}{\textbf{Model}} & \multirow{2}{*}{\textbf{Method}} & \multicolumn{2}{c}{\textbf{Alpaca}} & \multicolumn{2}{c}{\textbf{GSM8K}} & \multicolumn{2}{c}{\textbf{HumanEval}} & \multicolumn{2}{c}{\textbf{MT-Bench}} & \multicolumn{2}{c}{\textbf{QA}} & \multicolumn{2}{c}{\textbf{CNN/DM}} \\
\cmidrule(lr){3-4} \cmidrule(lr){5-6} \cmidrule(lr){7-8} \cmidrule(lr){9-10} \cmidrule(lr){11-12} \cmidrule(lr){13-14}
& & \textsc{Mat} & Spd. & \textsc{Mat} & Spd. & \textsc{Mat} & Spd. & \textsc{Mat} & Spd. & \textsc{Mat} & Spd. & \textsc{Mat} & Spd. \\

\midrule

\multirow{6}{*}{Vicuna-13B} & \eagle-3 & 6.61 & 3.78$\times$ & 6.74 & 3.87$\times$ & 8.31 & 4.77$\times$ & 7.12 & 4.05$\times$ & 5.24 & 3.03$\times$ & 6.93 & 3.43$\times$ \\
 & SD & 2.19 & 1.30$\times$ & 2.20 & 1.19$\times$ & 2.86 & 1.47$\times$ & 2.51 & 1.29$\times$ & 1.97 & 1.20$\times$ & 2.63 & 1.41$\times$ \\
 & \textsc{Medusa} & 2.44 & 1.90$\times$ & 2.63 & 2.05$\times$ & 2.78 & 2.18$\times$ & 2.58 & 2.01$\times$ & 2.10 & 1.62$\times$ & 2.09 & 1.56$\times$ \\
 & \textsc{Hydra} & 3.51 & 2.40$\times$ & 3.66 & 2.53$\times$ & 3.87 & 2.67$\times$ & 3.64 & 2.46$\times$ & 2.88 & 1.95$\times$ & 2.82 & 1.86$\times$ \\
 & OPT-Tree & 6.56 & 3.71$\times$ & 6.77 & 3.65$\times$ & 8.21 & 4.35$\times$ & 6.95 & 3.75$\times$ & 5.22 & 3.10$\times$ & 6.91 & 3.22$\times$ \\
 & \green \textbf{\method} & \green 6.36 & \green \textbf{3.94$\times$} & \green 6.80 & \green \textbf{3.99$\times$} & \green 9.48 & \green \textbf{5.16$\times$} & \green 7.29 & \green \textbf{4.27$\times$} & \green 5.03 & \green \textbf{3.26$\times$} & \green 7.20 & \green \textbf{3.58$\times$} \\

\midrule
\multirow{2}{*}{DSL-8B} & \eagle-3 & 5.64 & 3.23$\times$ & 7.40 & 4.24$\times$ & 6.70 & 3.85$\times$ & 5.82 & 3.33$\times$ & 5.02 & 2.86$\times$ & 5.03 & 2.89$\times$ \\
 & \green \textbf{\method} & \green 5.18 & \green \textbf{3.46$\times$} & \green 7.46 & \green \textbf{4.43$\times$} & \green 6.28 & \green \textbf{3.96$\times$} & \green 5.45 & \green \textbf{3.56$\times$} & \green 4.66 & \green \textbf{3.12$\times$} & \green 4.75 & \green \textbf{3.14$\times$} \\

\midrule

\multirow{2}{*}{Llama3-8B} & \eagle-3 & 6.93 & 3.92$\times$ & 6.42 & 3.63$\times$ & 7.08 & 4.05$\times$ & 6.38 & 3.67$\times$ & 5.36 & 3.01$\times$ & 5.46 & 3.06$\times$ \\
 & \green \textbf{\method} & \green 6.51 & \green \textbf{4.04$\times$} & \green 6.24 & \green \textbf{3.81$\times$} & \green 7.28 & \green \textbf{4.20$\times$} & \green 6.22 & \green \textbf{3.85$\times$} & \green 5.09 & \green \textbf{3.27$\times$} & \green 5.28 & \green \textbf{3.30$\times$} \\

\midrule

\multirow{2}{*}{Qwen3-8B} & \eagle-3 & 3.45 & 2.08$\times$ & 3.92 & 2.37$\times$ & 3.89 & 2.35$\times$ & 3.64 & 2.20$\times$ & 3.46 & 2.09$\times$ & 3.27 & 1.95$\times$ \\
 & \green \textbf{\method} & \green 3.39 & \green \textbf{2.44$\times$} & \green 3.85 & \green \textbf{2.67$\times$} & \green 3.78 & \green \textbf{2.69$\times$} & \green 3.60 & \green \textbf{2.57$\times$} & \green 3.41 & \green \textbf{2.46$\times$} & \green 3.20 & \green \textbf{2.30$\times$} \\

\midrule

\multirow{2}{*}{Qwen3-32B} & \eagle-3 & 2.75 & 1.83$\times$ & 3.36 & 2.22$\times$ & 2.98 & 1.96$\times$ & 2.98 & 1.90$\times$ & 2.66 & 1.78$\times$ & 2.57 & 1.57$\times$ \\
 & \green \textbf{\method} & \green 2.72 & \green \textbf{2.05$\times$} & \green 3.30 & \green \textbf{2.38$\times$} & \green 2.96 & \green \textbf{2.15$\times$} & \green 2.94 & \green \textbf{2.10$\times$} & \green 2.64 & \green \textbf{1.99$\times$} & \green 2.54 & \green \textbf{1.72$\times$} \\

\bottomrule
\end{tabular}%
}
\end{table*}

\subsection{Experimental Setup}

\paragraph{Tasks and Datasets.}
We evaluate \method on a comprehensive suite of LLM backbones, including Llama-3.1-8B-Instruct \citep{llama3}, Qwen3 8B and 32B \citep{qwen-3}, DeepSeek-R1-Distill-LLaMA-8B (DSL) \citep{deepseek-r1}, and Vicuna-13B \citep{mt-bench}.
Following standard protocols, we conduct evaluations on six diverse benchmarks: MT-Bench \citep{mt-bench}, Alpaca \citep{alpaca}, GSM8K \citep{gsm8k}, HumanEval \citep{humaneval}, CNN/DM \citep{cnndm}, and QA \citep{qa}, which are widely used benchmarks for instruction, math reasoning, code generation, chat, QA and summarization.

\paragraph{Implementation Details.}
We evaluate \method against the state-of-the-art tree-based method \eagle-3 \citep{eagle}. All experiments are conducted on a single NVIDIA H200 (140GB) GPU with batch size 1 by default.
The global token budget $N$ is set to \texttt{60} for both \method and \eagle-3.
The threshold $\mu$ in \method is set to \texttt{0.03} for all experiments.
Detailed configurations (hyperparameters, hardware, baselines, models) are provided in Appendix~\ref{app:implementation}.
We use two metrics: mean accepted tokens (\textsc{Mat}) and wall-time speedup (Spd.).

\subsection{Main Results}

We present the main comparison with \eagle-3 in Table~\ref{tab:main_results}.
We further evaluate \method under stochastic sampling ($T=1$) in Table~\ref{tab:temp_results}, and provide detailed ablation studies on tree initialization and budget settings in Appendix~\ref{app:ablation}.

We observe the following:
(a) \textbf{Universal Speedup.} \method consistently outperforms \eagle-3 across all 8 models and 6 datasets. Notably, it achieves up to $5.16\times$ speedup on HumanEval with Vicuna-13B and $2.30\times$ speedup on CNN/DM with Qwen3-8B, significantly surpassing the baseline's $4.77\times$ and $1.95\times$.
(b) \textbf{Reasoning Adaptability.} The performance gap is particularly pronounced in reasoning-intensive tasks. On GSM8K (Math) and HumanEval (Code), \method achieves substantial gains (e.g., $2.67\times$ vs $2.37\times$ on Qwen3-8B GSM8K). This verifies that our budget-driven adaptive expansion effectively captures correct paths in low-entropy reasoning steps where static trees often under-explore.
(c) \textbf{Robustness.} As shown in Table~\ref{tab:temp_results}, \method maintains strong performance with $T=1$. While static trees suffer from flattened distributions, \method's confidence-gated mechanism successfully adapts the tree topology, achieving $2.51\times$ speedup on Qwen3-8B (HumanEval) compared to \eagle-3's $2.20\times$.

\begin{figure}[t]
    \centering
    \includegraphics[width=\linewidth]{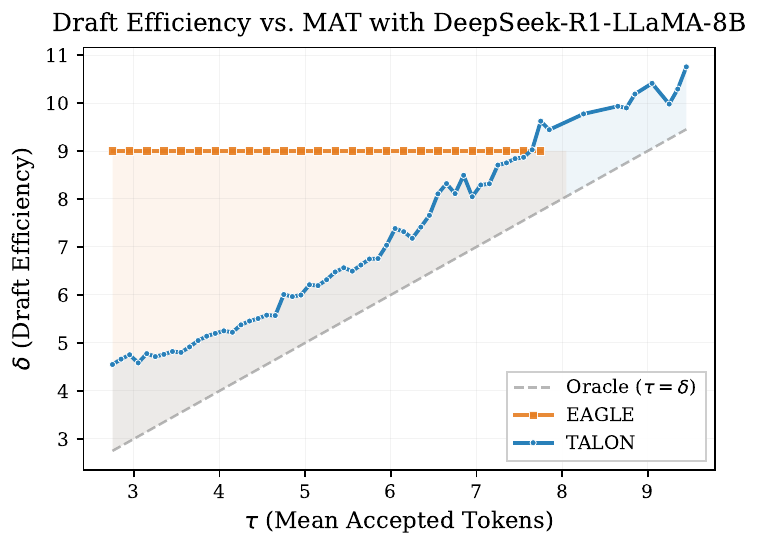}
    \caption{Draft Efficiency ($\delta$) vs. Mean Accepted Tokens ($\tau$). The \textcolor{gray}{gray dashed line} represents the Oracle baseline ($\tau = \delta$). The \textcolor{orange}{orange shaded region} highlights the computation waste of static methods (\eagle-3). \method (\textcolor{desc}{blue line and shaded region}) closely tracks the Oracle, minimizing waste by dynamically aligning draft cost with generation difficulty.}
    \label{fig:draft-effi-dsl-8b}
\end{figure}

\subsection{Evaluating \method with Draft Efficiency}

We evaluate the draft efficiency (defined in Section~\ref{sec:theoretical-analysis}) of \method and \eagle and visualize the relationship between the overhead ($\delta$) and the benefit ($\tau$) in Figure~\ref{fig:draft-effi-dsl-8b}. Static methods (orange line) pay a fixed high computational cost without considering context difficulty, resulting in sub-optimal speedup. In contrast, \method (blue line) closely approaches the zero-waste Oracle line ($\tau = \delta$). By dynamically shrinking the token trees in uncertain regions and expanding them in deterministic ones, \method effectively decouples draft cost from tree depth. This confirms that our framework maximizes speedup by ensuring computational resources are only invested where they yield high acceptance utility.
We also provide more visualization results of different models in Appendix~\ref{app:draft_efficiency}.

\subsection{Ablation Study}

\begin{figure*}[t]
    \centering
    \includegraphics[width=\textwidth]{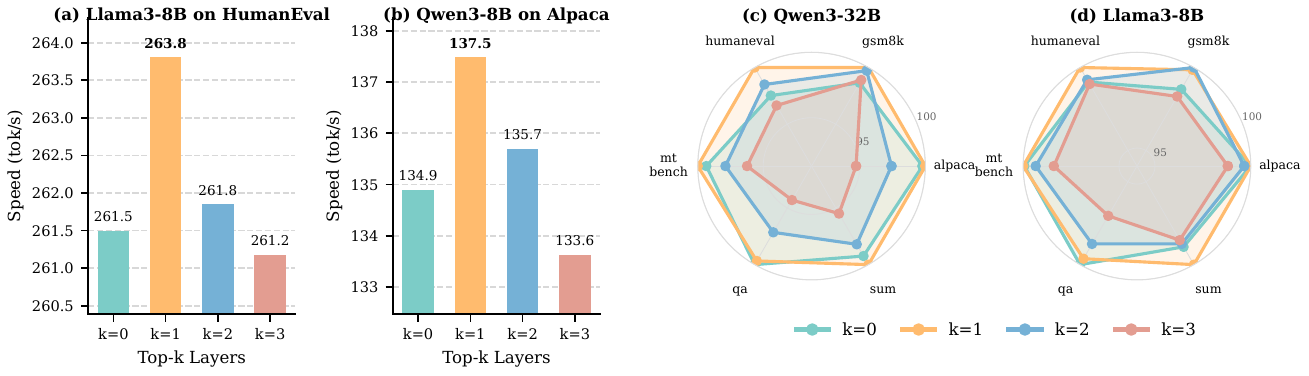}
    \caption{Ablation study on the number of initial Top-$K$ layers ($k$). $k=0$ represents the pure adaptive strategy without robust initialization, while $k \ge 1$ indicates using static Top-$K$ expansion for the first $k$ layers. (a)-(b) Wall-time speedup on HumanEval and Alpaca shows that $k=1$ (\method's default) achieves the highest throughput. (c)-(d) Radar charts across six benchmarks further confirm that $k=1$ (orange line) yields the most robust performance, outperforming both the uninitialized ($k=0$) and over-extended ($k \ge 2$) configurations.}
    \label{fig:ablation-topk-layer}
\end{figure*}

\subsubsection{Ablating Robust Tree Initialization}
\label{sssec:topk}

\method employs a hybrid strategy that begins with robust tree initialization for the first layer to ensure robustness, followed by confidence-gated expansion layers. However, a natural question arises, \textit{``Why is robust tree initialization necessary only for the first layer''?} To answer this question, we conduct an ablation study that using robust tree initialization for the first $k$ layers. $k=0$ corresponds to a pure confidence-gated strategy without robust tree initialization.

\paragraph{Necessity of Robust Initialization ($k=1 \ vs. \ k = 0$).} As shown in Fig.~\ref{fig:ablation-topk-layer} (a) and (b), removing the robust tree initialization leads to a noticeable drop in generation speed (e.g., declining from 263.8 to 261.5 tok/s on Llama3-8B).This degradation validates our hypothesis that draft models often suffer from root layer \textit{over-confidence}; without a forced Top-$K$ expansion at the first layer, the model risks falling into incorrect branches due to over-confidence, leading to early draft rejections.

\paragraph{Overhead of additional Robust Initialization ($k \ge 2$).} Conversely, extending the robust initialization phase to deeper layers ($k=2, 3$) yields suboptimal returns. It delays the transition to the more efficient confidence-gating expansion, thereby increasing the computational overhead ($\delta$) without a proportional gain in acceptance.The radar charts in Fig.~\ref{fig:ablation-topk-layer} (c) and (d) clearly illustrate that $k=1$ (orange line) consistently outperforms other configurations, confirming that a single layer of robust initialization followed immediately by adaptive gating achieves the optimal speedup.

\subsubsection{Ablating Confidence-gated Expansion}
\label{sssec:talon}
To further investigate the mechanism behind \method, we conduct two additional ablation studies on the influence of different threshold $\mu$ and different budget $N$ in Appendix~\ref{app:ablation}. The results show that the threshold $\mu$ in \method serves as a trade-off hyperparameter \textbf{between exploration and exploitation}. A larger $\mu$ leads to more deep-and-narrow draft trees, especially effective in reasoning tasks such as coding and math, while a smaller $\mu$ is more appropriate for some creative tasks. Moreover, the results demonstrate that \method is more flexible to different computation budget. For some resource-intensive scenarios, static methods waste more computation with fixed $K$ and $D$. However, \method seamlessly adapts to different computation budget by simply adjusting $N$.

\subsection{Case Study}
\label{ssec:case_study}

We also provide an real-world case of \method generated draft trees in Appendix~\ref{app:case_study}. Additionally, we provide an in-depth runtime breakdown of the tree construction phase in Appendix~\ref{app:tree-cons-overhead}, verifying that the algorithmic overhead of \method remains negligible even with large vocabularies.

\section{Conclusion}
\label{sec:conclusion}

In this work, we presented \method, a training-free framework that shifts speculative decoding from rigid geometric constraints to a flexible, budget-driven paradigm.
By employing a hybrid expansion strategy that combines robust initialization with confidence gating, \method dynamically shapes the draft tree—evolving into \textit{deep-and-narrow} chains for deterministic contexts or \textit{shallow-and-wide} branches for uncertain ones.
Extensive evaluations across 5 LLMs and 6 benchmarks demonstrate that \method consistently outperforms SOTA methods like \eagle-3, achieving up to $5.16\times$ speedup.

\newpage

\section{Limitations}
\label{sec:limitation}
While \method demonstrates significant speedups and adaptability across various benchmarks, we identify several limitations that present avenues for future research:

\paragraph{Scalability to Large Batch Sizes.} Our current evaluation focuses on latency-critical scenarios with a batch size of 1, which is the primary use case for real-time interaction. In high-throughput scenarios with large batch sizes, the compute-bound nature of the GPU may saturate, and the overhead of maintaining diverse dynamic tree structures for each request in a batch could become non-trivial. The memory management for varying tree topologies across a batch also presents implementation challenges. Extending budget-driven adaptive speculation to large-batch serving systems remains an open engineering challenge.

\paragraph{Hyperparameter Generalization.} While we found a fixed threshold ($\mu=0.03$) and budget ($N=60$) to be robust across most tested models and datasets, optimal performance in highly specialized domains might require task-specific tuning. Developing an auto-tuning mechanism that adjusts $\mu$ and $N$ on-the-fly based on acceptance history would be a valuable extension.

\section{Ethic Statements}
\label{sec:ethic}
\paragraph{Inheritance of Model Behaviors.} \method is an inference acceleration framework designed to speed up existing Large Language Models without modifying their weights. As a speculative decoding method, it aims to losslessly recover the distribution of the target model. Consequently, \method inherits the ethical properties, biases, and potential safety risks of the underlying target LLM and draft model. It does not introduce new capabilities for generating harmful content, nor does it mitigate existing biases in the base models. Users should continue to apply standard safety guardrails and alignment techniques to the target models deployed with \method.

\clearpage
\bibliography{reference}

@misc{gpt-5,
  title        = {GPT-5 System Card},
  author       = {{OpenAI Team}},
  year         = {2025},
  month        = aug,
  howpublished = {\url{https://cdn.openai.com/gpt-5-system-card.pdf}},
  note         = {System card, OpenAI},
}

@misc{gemini-3,
  title        = {Gemini 3 Pro Model Card},
  author       = {{Google DeepMind Team}},
  year         = {2025},
  month        = nov,
  howpublished = {\url{https://storage.googleapis.com/deepmind-media/Model-Cards/Gemini-3-Pro-Model-Card.pdf}},
  note         = {Model card, Google DeepMind}
}

@misc{opus-4-5,
  title        = {System Card: Claude Opus 4.5},
  author       = {{Anthropic Team}},
  year         = {2025},
  month        = nov,
  howpublished = {\url{https://assets.anthropic.com/m/64823ba7485345a7/Claude-Opus-4-5-System-Card.pdf}},
  note         = {System card, Anthropic}
}

@misc{qwen-3,
      title={Qwen3 Technical Report}, 
      author={QwenTeam and An Yang and Anfeng Li and Baosong Yang and Beichen Zhang and Binyuan Hui and Bo Zheng and Bowen Yu and Chang Gao and Chengen Huang and Chenxu Lv and Chujie Zheng and Dayiheng Liu and Fan Zhou and Fei Huang and Feng Hu and Hao Ge and Haoran Wei and Huan Lin and Jialong Tang and Jian Yang and Jianhong Tu and Jianwei Zhang and Jianxin Yang and Jiaxi Yang and Jing Zhou and Jingren Zhou and Junyang Lin and Kai Dang and Keqin Bao and Kexin Yang and Le Yu and Lianghao Deng and Mei Li and Mingfeng Xue and Mingze Li and Pei Zhang and Peng Wang and Qin Zhu and Rui Men and Ruize Gao and Shixuan Liu and Shuang Luo and Tianhao Li and Tianyi Tang and Wenbiao Yin and Xingzhang Ren and Xinyu Wang and Xinyu Zhang and Xuancheng Ren and Yang Fan and Yang Su and Yichang Zhang and Yinger Zhang and Yu Wan and Yuqiong Liu and Zekun Wang and Zeyu Cui and Zhenru Zhang and Zhipeng Zhou and Zihan Qiu},
      year={2025},
      eprint={2505.09388},
      archivePrefix={arXiv},
      primaryClass={cs.CL},
      url={https://arxiv.org/abs/2505.09388}, 
}

@misc{deepseek-r1,
      title={DeepSeek-R1: Incentivizing Reasoning Capability in LLMs via Reinforcement Learning}, 
      author={DeepSeek-AI and Daya Guo and Dejian Yang and Haowei Zhang and Junxiao Song and Ruoyu Zhang and Runxin Xu and Qihao Zhu and Shirong Ma and Peiyi Wang and Xiao Bi and Xiaokang Zhang and Xingkai Yu and Yu Wu and Z. F. Wu and Zhibin Gou and Zhihong Shao and Zhuoshu Li and Ziyi Gao and Aixin Liu and Bing Xue and Bingxuan Wang and Bochao Wu and Bei Feng and Chengda Lu and Chenggang Zhao and Chengqi Deng and Chenyu Zhang and Chong Ruan and Damai Dai and Deli Chen and Dongjie Ji and Erhang Li and Fangyun Lin and Fucong Dai and Fuli Luo and Guangbo Hao and Guanting Chen and Guowei Li and H. Zhang and Han Bao and Hanwei Xu and Haocheng Wang and Honghui Ding and Huajian Xin and Huazuo Gao and Hui Qu and Hui Li and Jianzhong Guo and Jiashi Li and Jiawei Wang and Jingchang Chen and Jingyang Yuan and Junjie Qiu and Junlong Li and J. L. Cai and Jiaqi Ni and Jian Liang and Jin Chen and Kai Dong and Kai Hu and Kaige Gao and Kang Guan and Kexin Huang and Kuai Yu and Lean Wang and Lecong Zhang and Liang Zhao and Litong Wang and Liyue Zhang and Lei Xu and Leyi Xia and Mingchuan Zhang and Minghua Zhang and Minghui Tang and Meng Li and Miaojun Wang and Mingming Li and Ning Tian and Panpan Huang and Peng Zhang and Qiancheng Wang and Qinyu Chen and Qiushi Du and Ruiqi Ge and Ruisong Zhang and Ruizhe Pan and Runji Wang and R. J. Chen and R. L. Jin and Ruyi Chen and Shanghao Lu and Shangyan Zhou and Shanhuang Chen and Shengfeng Ye and Shiyu Wang and Shuiping Yu and Shunfeng Zhou and Shuting Pan and S. S. Li and Shuang Zhou and Shaoqing Wu and Shengfeng Ye and Tao Yun and Tian Pei and Tianyu Sun and T. Wang and Wangding Zeng and Wanjia Zhao and Wen Liu and Wenfeng Liang and Wenjun Gao and Wenqin Yu and Wentao Zhang and W. L. Xiao and Wei An and Xiaodong Liu and Xiaohan Wang and Xiaokang Chen and Xiaotao Nie and Xin Cheng and Xin Liu and Xin Xie and Xingchao Liu and Xinyu Yang and Xinyuan Li and Xuecheng Su and Xuheng Lin and X. Q. Li and Xiangyue Jin and Xiaojin Shen and Xiaosha Chen and Xiaowen Sun and Xiaoxiang Wang and Xinnan Song and Xinyi Zhou and Xianzu Wang and Xinxia Shan and Y. K. Li and Y. Q. Wang and Y. X. Wei and Yang Zhang and Yanhong Xu and Yao Li and Yao Zhao and Yaofeng Sun and Yaohui Wang and Yi Yu and Yichao Zhang and Yifan Shi and Yiliang Xiong and Ying He and Yishi Piao and Yisong Wang and Yixuan Tan and Yiyang Ma and Yiyuan Liu and Yongqiang Guo and Yuan Ou and Yuduan Wang and Yue Gong and Yuheng Zou and Yujia He and Yunfan Xiong and Yuxiang Luo and Yuxiang You and Yuxuan Liu and Yuyang Zhou and Y. X. Zhu and Yanhong Xu and Yanping Huang and Yaohui Li and Yi Zheng and Yuchen Zhu and Yunxian Ma and Ying Tang and Yukun Zha and Yuting Yan and Z. Z. Ren and Zehui Ren and Zhangli Sha and Zhe Fu and Zhean Xu and Zhenda Xie and Zhengyan Zhang and Zhewen Hao and Zhicheng Ma and Zhigang Yan and Zhiyu Wu and Zihui Gu and Zijia Zhu and Zijun Liu and Zilin Li and Ziwei Xie and Ziyang Song and Zizheng Pan and Zhen Huang and Zhipeng Xu and Zhongyu Zhang and Zhen Zhang},
      year={2025},
      eprint={2501.12948},
      archivePrefix={arXiv},
      primaryClass={cs.CL},
      url={https://arxiv.org/abs/2501.12948}, 
}

@misc{glm,
      title={GLM-4.5: Agentic, Reasoning, and Coding (ARC) Foundation Models}, 
      author={GLM Team and Aohan Zeng and Xin Lv and Qinkai Zheng and Zhenyu Hou and Bin Chen and Chengxing Xie and Cunxiang Wang and Da Yin and Hao Zeng and Jiajie Zhang and Kedong Wang and Lucen Zhong and Mingdao Liu and Rui Lu and Shulin Cao and Xiaohan Zhang and Xuancheng Huang and Yao Wei and Yean Cheng and Yifan An and Yilin Niu and Yuanhao Wen and Yushi Bai and Zhengxiao Du and Zihan Wang and Zilin Zhu and Bohan Zhang and Bosi Wen and Bowen Wu and Bowen Xu and Can Huang and Casey Zhao and Changpeng Cai and Chao Yu and Chen Li and Chendi Ge and Chenghua Huang and Chenhui Zhang and Chenxi Xu and Chenzheng Zhu and Chuang Li and Congfeng Yin and Daoyan Lin and Dayong Yang and Dazhi Jiang and Ding Ai and Erle Zhu and Fei Wang and Gengzheng Pan and Guo Wang and Hailong Sun and Haitao Li and Haiyang Li and Haiyi Hu and Hanyu Zhang and Hao Peng and Hao Tai and Haoke Zhang and Haoran Wang and Haoyu Yang and He Liu and He Zhao and Hongwei Liu and Hongxi Yan and Huan Liu and Huilong Chen and Ji Li and Jiajing Zhao and Jiamin Ren and Jian Jiao and Jiani Zhao and Jianyang Yan and Jiaqi Wang and Jiayi Gui and Jiayue Zhao and Jie Liu and Jijie Li and Jing Li and Jing Lu and Jingsen Wang and Jingwei Yuan and Jingxuan Li and Jingzhao Du and Jinhua Du and Jinxin Liu and Junkai Zhi and Junli Gao and Ke Wang and Lekang Yang and Liang Xu and Lin Fan and Lindong Wu and Lintao Ding and Lu Wang and Man Zhang and Minghao Li and Minghuan Xu and Mingming Zhao and Mingshu Zhai and Pengfan Du and Qian Dong and Shangde Lei and Shangqing Tu and Shangtong Yang and Shaoyou Lu and Shijie Li and Shuang Li and Shuang-Li and Shuxun Yang and Sibo Yi and Tianshu Yu and Wei Tian and Weihan Wang and Wenbo Yu and Weng Lam Tam and Wenjie Liang and Wentao Liu and Xiao Wang and Xiaohan Jia and Xiaotao Gu and Xiaoying Ling and Xin Wang and Xing Fan and Xingru Pan and Xinyuan Zhang and Xinze Zhang and Xiuqing Fu and Xunkai Zhang and Yabo Xu and Yandong Wu and Yida Lu and Yidong Wang and Yilin Zhou and Yiming Pan and Ying Zhang and Yingli Wang and Yingru Li and Yinpei Su and Yipeng Geng and Yitong Zhu and Yongkun Yang and Yuhang Li and Yuhao Wu and Yujiang Li and Yunan Liu and Yunqing Wang and Yuntao Li and Yuxuan Zhang and Zezhen Liu and Zhen Yang and Zhengda Zhou and Zhongpei Qiao and Zhuoer Feng and Zhuorui Liu and Zichen Zhang and Zihan Wang and Zijun Yao and Zikang Wang and Ziqiang Liu and Ziwei Chai and Zixuan Li and Zuodong Zhao and Wenguang Chen and Jidong Zhai and Bin Xu and Minlie Huang and Hongning Wang and Juanzi Li and Yuxiao Dong and Jie Tang},
      year={2025},
      eprint={2508.06471},
      archivePrefix={arXiv},
      primaryClass={cs.CL},
      url={https://arxiv.org/abs/2508.06471}, 
}

@misc{mtp,
      title={Better \& Faster Large Language Models via Multi-token Prediction}, 
      author={Fabian Gloeckle and Badr Youbi Idrissi and Baptiste Rozière and David Lopez-Paz and Gabriel Synnaeve},
      year={2024},
      eprint={2404.19737},
      archivePrefix={arXiv},
      primaryClass={cs.CL},
      url={https://arxiv.org/abs/2404.19737}, 
}

@misc{palm-scaling,
      title={Efficiently Scaling Transformer Inference}, 
      author={Reiner Pope and Sholto Douglas and Aakanksha Chowdhery and Jacob Devlin and James Bradbury and Anselm Levskaya and Jonathan Heek and Kefan Xiao and Shivani Agrawal and Jeff Dean},
      year={2022},
      eprint={2211.05102},
      archivePrefix={arXiv},
      primaryClass={cs.LG},
      url={https://arxiv.org/abs/2211.05102}, 
}

@InProceedings{sps1,
  title = 	 {Fast Inference from Transformers via Speculative Decoding},
  author =       {Leviathan, Yaniv and Kalman, Matan and Matias, Yossi},
  booktitle = 	 {Proceedings of the 40th International Conference on Machine Learning},
  pages = 	 {19274--19286},
  year = 	 {2023},
  editor = 	 {Krause, Andreas and Brunskill, Emma and Cho, Kyunghyun and Engelhardt, Barbara and Sabato, Sivan and Scarlett, Jonathan},
  volume = 	 {202},
  series = 	 {Proceedings of Machine Learning Research},
  month = 	 {23--29 Jul},
  publisher =    {PMLR},
  url = 	 {https://proceedings.mlr.press/v202/leviathan23a.html}
}

@misc{sps2,
      title={Accelerating Large Language Model Decoding with Speculative Sampling}, 
      author={Charlie Chen and Sebastian Borgeaud and Geoffrey Irving and Jean-Baptiste Lespiau and Laurent Sifre and John Jumper},
      year={2023},
      eprint={2302.01318},
      archivePrefix={arXiv},
      primaryClass={cs.CL},
      url={https://arxiv.org/abs/2302.01318}, 
}

@inproceedings{
pearl,
title={{PEARL}: Parallel Speculative Decoding with Adaptive Draft Length},
author={Tianyu Liu and Yun Li and Qitan Lv and Kai Liu and Jianchen Zhu and Winston Hu and Xiao Sun},
booktitle={The Thirteenth International Conference on Learning Representations},
year={2025},
url={https://openreview.net/forum?id=QOXrVMiHGK}
}

@inproceedings{
specinfer,
author = {Miao, Xupeng and Oliaro, Gabriele and Zhang, Zhihao and Cheng, Xinhao and Wang, Zeyu and Zhang, Zhengxin and Wong, Rae Ying Yee and Zhu, Alan and Yang, Lijie and Shi, Xiaoxiang and Shi, Chunan and Chen, Zhuoming and Arfeen, Daiyaan and Abhyankar, Reyna and Jia, Zhihao},
title = {SpecInfer: Accelerating Large Language Model Serving with Tree-based Speculative Inference and Verification},
year = {2024},
isbn = {9798400703867},
publisher = {Association for Computing Machinery},
address = {New York, NY, USA},
url = {https://doi.org/10.1145/3620666.3651335},
doi = {10.1145/3620666.3651335},
booktitle = {Proceedings of the 29th ACM International Conference on Architectural Support for Programming Languages and Operating Systems, Volume 3},
pages = {932–949},
numpages = {18},
location = {La Jolla, CA, USA},
series = {ASPLOS '24}
}

@inproceedings{medusa,
author = {Cai, Tianle and Li, Yuhong and Geng, Zhengyang and Peng, Hongwu and Lee, Jason D. and Chen, Deming and Dao, Tri},
title = {MEDUSA: Simple LLM inference acceleration framework with multiple decoding heads},
year = {2024},
publisher = {JMLR.org},
booktitle = {Proceedings of the 41st International Conference on Machine Learning},
articleno = {203},
numpages = {27},
location = {Vienna, Austria},
series = {ICML'24}
}

@misc{eagle,
      title={EAGLE: Speculative Sampling Requires Rethinking Feature Uncertainty}, 
      author={Yuhui Li and Fangyun Wei and Chao Zhang and Hongyang Zhang},
      year={2025},
      eprint={2401.15077},
      archivePrefix={arXiv},
      primaryClass={cs.LG},
      url={https://arxiv.org/abs/2401.15077}, 
}

@inproceedings{eagle2,
    title = "{EAGLE}-2: Faster Inference of Language Models with Dynamic Draft Trees",
    author = "Li, Yuhui  and
      Wei, Fangyun  and
      Zhang, Chao  and
      Zhang, Hongyang",
    editor = "Al-Onaizan, Yaser  and
      Bansal, Mohit  and
      Chen, Yun-Nung",
    booktitle = "Proceedings of the 2024 Conference on Empirical Methods in Natural Language Processing",
    month = nov,
    year = "2024",
    address = "Miami, Florida, USA",
    publisher = "Association for Computational Linguistics",
    url = "https://aclanthology.org/2024.emnlp-main.422/",
    doi = "10.18653/v1/2024.emnlp-main.422",
    pages = "7421--7432"
}

@misc{eagle3,
      title={EAGLE-3: Scaling up Inference Acceleration of Large Language Models via Training-Time Test}, 
      author={Yuhui Li and Fangyun Wei and Chao Zhang and Hongyang Zhang},
      year={2025},
      eprint={2503.01840},
      archivePrefix={arXiv},
      primaryClass={cs.CL},
      url={https://arxiv.org/abs/2503.01840}, 
}

@misc{opt-tree,
      title={OPT-Tree: Speculative Decoding with Adaptive Draft Tree Structure}, 
      author={Jikai Wang and Yi Su and Juntao Li and Qingrong Xia and Zi Ye and Xinyu Duan and Zhefeng Wang and Min Zhang},
      year={2024},
      eprint={2406.17276},
      archivePrefix={arXiv},
      primaryClass={cs.CL},
      url={https://arxiv.org/abs/2406.17276}, 
}

@inproceedings{lade,
author = {Fu, Yichao and Bailis, Peter and Stoica, Ion and Zhang, Hao},
title = {Break the sequential dependency of LLM inference using LOOKAHEAD DECODING},
year = {2024},
publisher = {JMLR.org},
booktitle = {Proceedings of the 41st International Conference on Machine Learning},
articleno = {561},
numpages = {20},
location = {Vienna, Austria},
series = {ICML'24}
}

@misc{flash-tree-attn,
      title={DeFT: Decoding with Flash Tree-attention for Efficient Tree-structured LLM Inference}, 
      author={Jinwei Yao and Kaiqi Chen and Kexun Zhang and Jiaxuan You and Binhang Yuan and Zeke Wang and Tao Lin},
      year={2025},
      eprint={2404.00242},
      archivePrefix={arXiv},
      primaryClass={cs.CL},
      url={https://arxiv.org/abs/2404.00242}, 
}

@inproceedings{mt-bench,
title={Judging {LLM}-as-a-Judge with {MT}-Bench and Chatbot Arena},
author={Lianmin Zheng and Wei-Lin Chiang and Ying Sheng and Siyuan Zhuang and Zhanghao Wu and Yonghao Zhuang and Zi Lin and Zhuohan Li and Dacheng Li and Eric Xing and Hao Zhang and Joseph E. Gonzalez and Ion Stoica},
booktitle={Thirty-seventh Conference on Neural Information Processing Systems Datasets and Benchmarks Track},
year={2023},
url={https://openreview.net/forum?id=uccHPGDlao}
}

@inproceedings{glide,
author = {Du, Cunxiao and Jiang, Jing and Yuanchen, Xu and Wu, Jiawei and Yu, Sicheng and Li, Yongqi and Li, Shenggui and Xu, Kai and Nie, Liqiang and Tu, Zhaopeng and You, Yang},
title = {GLIDE with a CAPE: a low-hassle method to accelerate speculative decoding},
year = {2024},
publisher = {JMLR.org},
booktitle = {Proceedings of the 41st International Conference on Machine Learning},
articleno = {465},
numpages = {17},
location = {Vienna, Austria},
series = {ICML'24}
}

@inproceedings{
minp,
title={Turning Up the Heat: Min-p Sampling for Creative and Coherent {LLM} Outputs},
author={Nguyen Nhat Minh and Andrew Baker and Clement Neo and Allen G Roush and Andreas Kirsch and Ravid Shwartz-Ziv},
booktitle={The Thirteenth International Conference on Learning Representations},
year={2025},
url={https://openreview.net/forum?id=FBkpCyujtS}
}

@misc{llama3,
      title={The Llama 3 Herd of Models}, 
      author={Llama Team},
      year={2024},
      eprint={2407.21783},
      archivePrefix={arXiv},
      primaryClass={cs.AI},
      url={https://arxiv.org/abs/2407.21783}, 
}

@misc{alpaca,
      title={Enhancing Chat Language Models by Scaling High-quality Instructional Conversations}, 
      author={Ning Ding and Yulin Chen and Bokai Xu and Yujia Qin and Zhi Zheng and Shengding Hu and Zhiyuan Liu and Maosong Sun and Bowen Zhou},
      year={2023},
      eprint={2305.14233},
      archivePrefix={arXiv},
      primaryClass={cs.CL},
      url={https://arxiv.org/abs/2305.14233}, 
}

@misc{humaneval,
      title={Evaluating Large Language Models Trained on Code}, 
      author={Mark Chen and Jerry Tworek and Heewoo Jun and Qiming Yuan and Henrique Ponde de Oliveira Pinto and Jared Kaplan and Harri Edwards and Yuri Burda and Nicholas Joseph and Greg Brockman and Alex Ray and Raul Puri and Gretchen Krueger and Michael Petrov and Heidy Khlaaf and Girish Sastry and Pamela Mishkin and Brooke Chan and Scott Gray and Nick Ryder and Mikhail Pavlov and Alethea Power and Lukasz Kaiser and Mohammad Bavarian and Clemens Winter and Philippe Tillet and Felipe Petroski Such and Dave Cummings and Matthias Plappert and Fotios Chantzis and Elizabeth Barnes and Ariel Herbert-Voss and William Hebgen Guss and Alex Nichol and Alex Paino and Nikolas Tezak and Jie Tang and Igor Babuschkin and Suchir Balaji and Shantanu Jain and William Saunders and Christopher Hesse and Andrew N. Carr and Jan Leike and Josh Achiam and Vedant Misra and Evan Morikawa and Alec Radford and Matthew Knight and Miles Brundage and Mira Murati and Katie Mayer and Peter Welinder and Bob McGrew and Dario Amodei and Sam McCandlish and Ilya Sutskever and Wojciech Zaremba},
      year={2021},
      eprint={2107.03374},
      archivePrefix={arXiv},
      primaryClass={cs.LG},
      url={https://arxiv.org/abs/2107.03374}, 
}

@misc{gsm8k,
      title={Training Verifiers to Solve Math Word Problems}, 
      author={Karl Cobbe and Vineet Kosaraju and Mohammad Bavarian and Mark Chen and Heewoo Jun and Lukasz Kaiser and Matthias Plappert and Jerry Tworek and Jacob Hilton and Reiichiro Nakano and Christopher Hesse and John Schulman},
      year={2021},
      eprint={2110.14168},
      archivePrefix={arXiv},
      primaryClass={cs.LG},
      url={https://arxiv.org/abs/2110.14168}, 
}

@inproceedings{cnndm,
    title = "Abstractive Text Summarization using Sequence-to-sequence {RNN}s and Beyond",
    author = "Nallapati, Ramesh  and
      Zhou, Bowen  and
      dos Santos, Cicero  and
      Gu{\ensuremath{\dot{}}}l{\c{c}}ehre, {\c{C}}a{\u{g}}lar  and
      Xiang, Bing",
    editor = "Riezler, Stefan  and
      Goldberg, Yoav",
    booktitle = "Proceedings of the 20th {SIGNLL} Conference on Computational Natural Language Learning",
    month = aug,
    year = "2016",
    address = "Berlin, Germany",
    publisher = "Association for Computational Linguistics",
    url = "https://aclanthology.org/K16-1028/",
    doi = "10.18653/v1/K16-1028",
    pages = "280--290"
}

@article{qa,
    author = {Kwiatkowski, Tom and Palomaki, Jennimaria and Redfield, Olivia and Collins, Michael and Parikh, Ankur and Alberti, Chris and Epstein, Danielle and Polosukhin, Illia and Devlin, Jacob and Lee, Kenton and Toutanova, Kristina and Jones, Llion and Kelcey, Matthew and Chang, Ming-Wei and Dai, Andrew
                        M. and Uszkoreit, Jakob and Le, Quoc and Petrov, Slav},
    title = {Natural Questions: A Benchmark for Question Answering
                    Research},
    journal = {Transactions of the Association for Computational Linguistics},
    volume = {7},
    pages = {453-466},
    year = {2019},
    month = {08},
    issn = {2307-387X},
    doi = {10.1162/tacl_a_00276},
    url = {https://doi.org/10.1162/tacl_a_00276},
    eprint = {https://direct.mit.edu/tacl/article-pdf/doi/10.1162/tacl_a_00276/1923288/tacl_a_00276.pdf},
}

@inproceedings{draft_and_verify,
    title = "Draft {\&} Verify: Lossless Large Language Model Acceleration via Self-Speculative Decoding",
    author = "Zhang, Jun  and
      Wang, Jue  and
      Li, Huan  and
      Shou, Lidan  and
      Chen, Ke  and
      Chen, Gang  and
      Mehrotra, Sharad",
    editor = "Ku, Lun-Wei  and
      Martins, Andre  and
      Srikumar, Vivek",
    booktitle = "Proceedings of the 62nd Annual Meeting of the Association for Computational Linguistics (Volume 1: Long Papers)",
    month = aug,
    year = "2024",
    address = "Bangkok, Thailand",
    publisher = "Association for Computational Linguistics",
    url = "https://aclanthology.org/2024.acl-long.607/",
    doi = "10.18653/v1/2024.acl-long.607",
    pages = "11263--11282"
}

@inproceedings{rest,
    title = "{REST}: Retrieval-Based Speculative Decoding",
    author = "He, Zhenyu  and
      Zhong, Zexuan  and
      Cai, Tianle  and
      Lee, Jason  and
      He, Di",
    editor = "Duh, Kevin  and
      Gomez, Helena  and
      Bethard, Steven",
    booktitle = "Proceedings of the 2024 Conference of the North American Chapter of the Association for Computational Linguistics: Human Language Technologies (Volume 1: Long Papers)",
    month = jun,
    year = "2024",
    address = "Mexico City, Mexico",
    publisher = "Association for Computational Linguistics",
    url = "https://aclanthology.org/2024.naacl-long.88/",
    doi = "10.18653/v1/2024.naacl-long.88",
    pages = "1582--1595"
}

@misc{pld,
    title = {Prompt Lookup Decoding},
    author = {Apoorv Saxena},
    year = {2023},
    month = {November},
    url = {https://github.com/apoorvumang/prompt-lookup-decoding/}
}

@inproceedings{
block_verification,
title={Block Verification Accelerates Speculative Decoding},
author={Ziteng Sun and Uri Mendlovic and Yaniv Leviathan and Asaf Aharoni and Jae Hun Ro and Ahmad Beirami and Ananda Theertha Suresh},
booktitle={The Thirteenth International Conference on Learning Representations},
year={2025},
url={https://openreview.net/forum?id=frsg32u0rO}
}

@misc{specdec++,
      title={SpecDec++: Boosting Speculative Decoding via Adaptive Candidate Lengths}, 
      author={Kaixuan Huang and Xudong Guo and Mengdi Wang},
      year={2025},
      eprint={2405.19715},
      archivePrefix={arXiv},
      primaryClass={cs.CL},
      url={https://arxiv.org/abs/2405.19715}, 
}

@misc{adaeagle,
      title={AdaEAGLE: Optimizing Speculative Decoding via Explicit Modeling of Adaptive Draft Structures}, 
      author={Situo Zhang and Hankun Wang and Da Ma and Zichen Zhu and Lu Chen and Kunyao Lan and Kai Yu},
      year={2024},
      eprint={2412.18910},
      archivePrefix={arXiv},
      primaryClass={cs.AI},
      url={https://arxiv.org/abs/2412.18910}, 
}

@misc{dyspec,
      title={DySpec: Faster Speculative Decoding with Dynamic Token Tree Structure}, 
      author={Yunfan Xiong and Ruoyu Zhang and Yanzeng Li and Tianhao Wu and Lei Zou},
      year={2024},
      eprint={2410.11744},
      archivePrefix={arXiv},
      primaryClass={cs.LG},
      url={https://arxiv.org/abs/2410.11744}, 
}

@misc{sequoia,
      title={Sequoia: Scalable, Robust, and Hardware-aware Speculative Decoding}, 
      author={Zhuoming Chen and Avner May and Ruslan Svirschevski and Yuhsun Huang and Max Ryabinin and Zhihao Jia and Beidi Chen},
      year={2025},
      eprint={2402.12374},
      archivePrefix={arXiv},
      primaryClass={cs.CL},
      url={https://arxiv.org/abs/2402.12374}, 
}

@inproceedings{falcon,
author = {Gao, Xiangxiang and Xie, Weisheng and Xiang, Yiwei and Ji, Feng},
title = {Falcon: faster and parallel inference of large language models through enhanced semi-autoregressive drafting and custom-designed decoding tree},
year = {2025},
isbn = {978-1-57735-897-8},
publisher = {AAAI Press},
url = {https://doi.org/10.1609/aaai.v39i22.34566},
doi = {10.1609/aaai.v39i22.34566},
booktitle = {Proceedings of the Thirty-Ninth AAAI Conference on Artificial Intelligence and Thirty-Seventh Conference on Innovative Applications of Artificial Intelligence and Fifteenth Symposium on Educational Advances in Artificial Intelligence},
articleno = {2668},
numpages = {9},
series = {AAAI'25/IAAI'25/EAAI'25}
}

@misc{c2t,
      title={C2T: A Classifier-Based Tree Construction Method in Speculative Decoding}, 
      author={Feiye Huo and Jianchao Tan and Kefeng Zhang and Xunliang Cai and Shengli Sun},
      year={2025},
      eprint={2502.13652},
      archivePrefix={arXiv},
      primaryClass={cs.CL},
      url={https://arxiv.org/abs/2502.13652}, 
}

@misc{hydra,
      title={Hydra: Sequentially-Dependent Draft Heads for Medusa Decoding}, 
      author={Zachary Ankner and Rishab Parthasarathy and Aniruddha Nrusimha and Christopher Rinard and Jonathan Ragan-Kelley and William Brandon},
      year={2024},
      eprint={2402.05109},
      archivePrefix={arXiv},
      primaryClass={cs.LG},
      url={https://arxiv.org/abs/2402.05109}, 
}

@misc{clover,
      title={Clover: Regressive Lightweight Speculative Decoding with Sequential Knowledge}, 
      author={Bin Xiao and Chunan Shi and Xiaonan Nie and Fan Yang and Xiangwei Deng and Lei Su and Weipeng Chen and Bin Cui},
      year={2024},
      eprint={2405.00263},
      archivePrefix={arXiv},
      primaryClass={cs.CL},
      url={https://arxiv.org/abs/2405.00263}, 
}

@misc{eesd,
      title={Speculative Decoding via Early-exiting for Faster LLM Inference with Thompson Sampling Control Mechanism}, 
      author={Jiahao Liu and Qifan Wang and Jingang Wang and Xunliang Cai},
      year={2024},
      eprint={2406.03853},
      archivePrefix={arXiv},
      primaryClass={cs.CL},
      url={https://arxiv.org/abs/2406.03853}, 
}

@misc{tokenrecycling,
      title={Turning Trash into Treasure: Accelerating Inference of Large Language Models with Token Recycling}, 
      author={Xianzhen Luo and Yixuan Wang and Qingfu Zhu and Zhiming Zhang and Xuanyu Zhang and Qing Yang and Dongliang Xu and Wanxiang Che},
      year={2024},
      eprint={2408.08696},
      archivePrefix={arXiv},
      primaryClass={cs.CL},
      url={https://arxiv.org/abs/2408.08696}, 
}

@misc{samdecoding,
      title={SAM Decoding: Speculative Decoding via Suffix Automaton}, 
      author={Yuxuan Hu and Ke Wang and Xiaokang Zhang and Fanjin Zhang and Cuiping Li and Hong Chen and Jing Zhang},
      year={2024},
      eprint={2411.10666},
      archivePrefix={arXiv},
      primaryClass={cs.CL},
      url={https://arxiv.org/abs/2411.10666}, 
}

@misc{judge,
      title={Judge Decoding: Faster Speculative Sampling Requires Going Beyond Model Alignment}, 
      author={Gregor Bachmann and Sotiris Anagnostidis and Albert Pumarola and Markos Georgopoulos and Artsiom Sanakoyeu and Yuming Du and Edgar Schönfeld and Ali Thabet and Jonas Kohler},
      year={2025},
      eprint={2501.19309},
      archivePrefix={arXiv},
      primaryClass={cs.LG},
      url={https://arxiv.org/abs/2501.19309}, 
}

@misc{coral,
      title={CORAL: Learning Consistent Representations across Multi-step Training with Lighter Speculative Drafter}, 
      author={Yepeng Weng and Dianwen Mei and Huishi Qiu and Xujie Chen and Li Liu and Jiang Tian and Zhongchao Shi},
      year={2025},
      eprint={2502.16880},
      archivePrefix={arXiv},
      primaryClass={cs.CL},
      url={https://arxiv.org/abs/2502.16880}, 
}

@inproceedings{
hass,
title={Learning Harmonized Representations for Speculative Sampling},
author={Lefan Zhang and Xiaodan Wang and Yanhua Huang and Ruiwen Xu},
booktitle={The Thirteenth International Conference on Learning Representations},
year={2025},
url={https://openreview.net/forum?id=T9u56s7mbk}
}

@misc{disco,
      title={Dynamic Speculation Lookahead Accelerates Speculative Decoding of Large Language Models}, 
      author={Jonathan Mamou and Oren Pereg and Daniel Korat and Moshe Berchansky and Nadav Timor and Moshe Wasserblat and Roy Schwartz},
      year={2024},
      eprint={2405.04304},
      archivePrefix={arXiv},
      primaryClass={cs.CL},
      url={https://arxiv.org/abs/2405.04304}, 
}

@misc{chimera,
      title={Chimera: A Lossless Decoding Method for Accelerating Large Language Models Inference by Fusing all Tokens}, 
      author={Ziqian Zeng and Jiahong Yu and Qianshi Pang and Zihao Wang and Huiping Zhuang and Hongen Shao and Xiaofeng Zou},
      year={2024},
      eprint={2402.15758},
      archivePrefix={arXiv},
      primaryClass={cs.CL},
      url={https://arxiv.org/abs/2402.15758}, 
}

@misc{logitspec,
      title={LogitSpec: Accelerating Retrieval-based Speculative Decoding via Next Next Token Speculation}, 
      author={Tianyu Liu and Qitan Lv and Hao Li and Xing Gao and Xiao Sun},
      year={2025},
      eprint={2507.01449},
      archivePrefix={arXiv},
      primaryClass={cs.CL},
      url={https://arxiv.org/abs/2507.01449}, 
}

@inproceedings{transformers,
    title = "Transformers: State-of-the-Art Natural Language Processing",
    author = "Thomas Wolf and Lysandre Debut and Victor Sanh and Julien Chaumond and Clement Delangue and Anthony Moi and Pierric Cistac and Tim Rault and Rémi Louf and Morgan Funtowicz and Joe Davison and Sam Shleifer and Patrick von Platen and Clara Ma and Yacine Jernite and Julien Plu and Canwen Xu and Teven Le Scao and Sylvain Gugger and Mariama Drame and Quentin Lhoest and Alexander M. Rush",
    booktitle = "Proceedings of the 2020 Conference on Empirical Methods in Natural Language Processing: System Demonstrations",
    month = oct,
    year = "2020",
    address = "Online",
    publisher = "Association for Computational Linguistics",
    url = "https://www.aclweb.org/anthology/2020.emnlp-demos.6",
    pages = "38--45"
}

@misc{pytorch,
      title={PyTorch: An Imperative Style, High-Performance Deep Learning Library}, 
      author={Adam Paszke and Sam Gross and Francisco Massa and Adam Lerer and James Bradbury and Gregory Chanan and Trevor Killeen and Zeming Lin and Natalia Gimelshein and Luca Antiga and Alban Desmaison and Andreas Köpf and Edward Yang and Zach DeVito and Martin Raison and Alykhan Tejani and Sasank Chilamkurthy and Benoit Steiner and Lu Fang and Junjie Bai and Soumith Chintala},
      year={2019},
      eprint={1912.01703},
      archivePrefix={arXiv},
      primaryClass={cs.LG},
      url={https://arxiv.org/abs/1912.01703}, 
}

@misc{cuda,
  author={NVIDIA and Vingelmann, Péter and Fitzek, Frank H.P.},
  title={CUDA, release: 10.2.89},
  year={2020},
  url={https://developer.nvidia.com/cuda-toolkit},
}

@inproceedings{
zipf,
title={The Curious Case of Neural Text Degeneration},
author={Ari Holtzman and Jan Buys and Li Du and Maxwell Forbes and Yejin Choi},
booktitle={International Conference on Learning Representations},
year={2020},
url={https://openreview.net/forum?id=rygGQyrFvH}
}

\newpage
\appendix
\section{Formalized Algorithms}
\label{app:alg}

\begin{algorithm}[t]
    \caption{Static Tree Construction (\eagle)}
    \label{alg:eagle}
    \begin{algorithmic}[1]
        \REQUIRE Draft Model $\mathcal{M}_d$, Prefix $x_{\le t}$, Depth $D$, Width $K$, Budget $N$
        \STATE $\mathcal{T} \leftarrow \{x_t\}$, $\mathcal{P}_0 \leftarrow \{x_t\}$, $p(x_t) \leftarrow 1.0$
        \STATE \textcolor{rej}{Fixed depth and width for-loop}
        \FOR{\textcolor{rej}{$d = 0$ to $D$}}
            \STATE $\mathcal{S}_d \leftarrow \emptyset$
            \STATE \textcolor{rej}{$\triangleright$ Expand to $K\times K$ child nodes}
            \FOR{$v \in \mathcal{P}_d$}
                \STATE $C_v \leftarrow \text{Top-}K(P_{\mathcal{M}_d}(\cdot \mid x_{\le v}))$
                \STATE $p(u) \leftarrow p(v) \cdot P_{\mathcal{M}_d}(u \mid x_{\le v}), \forall u \in C_v$
                \STATE $\mathcal{S}_d \leftarrow \mathcal{S}_d \cup C_v$
            \ENDFOR
            \STATE \textcolor{rej}{$\triangleright$ Shrink to $K$ next-layer parent nodes}
            \STATE $\mathcal{P}_{d+1} \leftarrow \text{Top-}K(\mathcal{S}_d \text{ by } p(u))$
            \STATE $\mathcal{T} \leftarrow \mathcal{T} \cup \mathcal{P}_{d+1}$
        \ENDFOR
        \STATE \textcolor{rej}{$\triangleright$ Final pruning to budget $N$}
        \STATE $\mathcal{T} \leftarrow \text{Top-}N(\mathcal{T} \text{ by } p(u))$
        \RETURN $\mathcal{T}$
    \end{algorithmic}
\end{algorithm}

\begin{algorithm}[t]
    \caption{Adaptive Tree Construction (\method)}
    \label{alg:talon}
    \begin{algorithmic}[1]
        \REQUIRE Draft Model $\mathcal{M}_d$, Prefix $x_{\le t}$, Budget $N$, Width $K$, Threshold $\mu$
        \STATE $\mathcal{T} \leftarrow \{x_t\}$, $\mathcal{P}_0 \leftarrow \{x_t\}$, $p(x_t) \leftarrow 1.0$
        \STATE $d \leftarrow 0$
        \STATE \textcolor{desc}{$\triangleright$ Budget-driven adaptive tree expansion}
        \WHILE{\textcolor{desc}{$|\mathcal{T}| < N$}}
            \STATE $\mathcal{S}_d \leftarrow \emptyset$
            \STATE \textcolor{desc}{$\triangleright$ Gather next-layer candidate set $\mathcal{S}_d$}
            \FOR{$v \in \mathcal{P}_d$}
                \STATE $C_v \leftarrow \{(v, w) \mid w \in \mathcal{V}\}$
                \STATE $p(u) \leftarrow p(v) \cdot P_{\mathcal{M}_d}(w \mid x_{\le v}), \forall u=(v, w) \in C_v$
                \STATE $\mathcal{S}_d \leftarrow \mathcal{S}_d \cup C_v$
            \ENDFOR
            
            \STATE \textcolor{desc}{$\triangleright$ Hybrid Expansion Strategy}
            \IF{$d=0$}
                \STATE \textcolor{acc}{$\triangleright$ Robust Init}
                \STATE $\mathcal{P}_{d+1} \leftarrow \text{Top-}K(\mathcal{S}_d \text{ by } p(u))$
            \ELSE
                \STATE \textcolor{acc}{$\triangleright$ Confidence Gating}
                \STATE $m_d \leftarrow \max_{u \in \mathcal{S}_d} p(u)$
                \STATE $\mathcal{P}_{d+1} \leftarrow \{u \in \mathcal{S}_d \mid p(u) \ge \mu \cdot m_d\}$
            \ENDIF
            
            \STATE \textcolor{desc}{$\triangleright$ Budget Check}
            \IF{$|\mathcal{T}| + |\mathcal{P}_{d+1}| > N$}
                \STATE $\mathcal{P}_{d+1} \leftarrow \text{Top-}(N - |\mathcal{T}|)(\mathcal{P}_{d+1} \text{ by } p(u))$
            \ENDIF
            
            \STATE $\mathcal{T} \leftarrow \mathcal{T} \cup \mathcal{P}_{d+1}$
            \STATE $d \leftarrow d+1$
        \ENDWHILE
        \RETURN $\mathcal{T}$
    \end{algorithmic}
\end{algorithm}

\begin{figure*}[htbp]
    \centering
    \includegraphics[width=\textwidth]{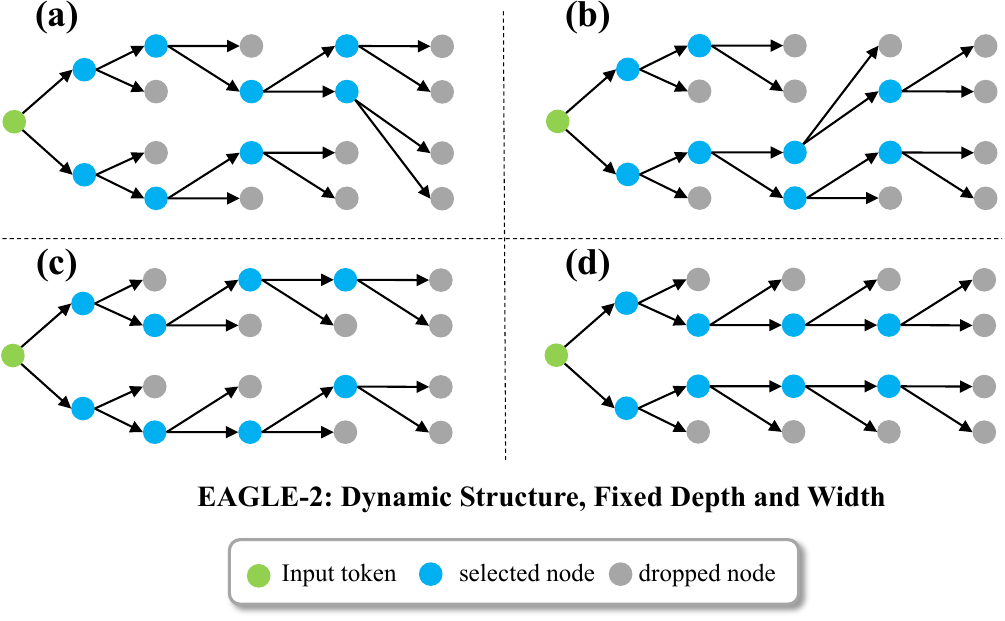}
    \caption{Examples of \eagle-style draft token trees. At each layer, \eagle uses a top-$K$ operation to select $K$ nodes as next layer parents. In this way, \eagle can dynamically adjust the edges between last-layer parents and current-layer children. However, they cannot adjust $K$ to adapt for contexts, which leads to significant resource wastes.}
    \label{fig:eagle2_examples}
\end{figure*}

In this section, we provide the formalized algorithms for both the static baseline and our proposed framework to illustrate the structural differences in their expansion strategies.

Algorithm \ref{alg:eagle} outlines the static tree construction strategy employed by state-of-the-art methods such as \eagle. This approach relies on rigid geometric constraints defined by a fixed depth $D$ and width $K$. The construction proceeds layer-by-layer up to depth $D$. At each step $d$, the draft model expands $K \times K$ child nodes from the parent set $\mathcal{P}_d$ and subsequently applies a Top-K operation to shrink the candidates back to size $K$ for the next layer. Finally, the tree is pruned globally to meet the token budget $N$. This "expand-then-shrink" mechanism enforces a static topology regardless of generation difficulty, often generating redundant nodes that are discarded during intermediate steps. Note that ``static'' here means that each layer of the draft tree has a fixed number of nodes regardless of the context difficulty. The ``dynamic'' claimed in \citep{eagle2} means that the draft model dynamically select top-$K$ nodes as next-layer parents, but it cannot adjust $K$ to adapt for context. Figure~\ref{fig:eagle2_examples} shows 4 different \eagle-style draft trees with static tree topology.

Algorithm \ref{alg:talon} presents the details of \method, our proposed budget-driven adaptive framework. Unlike static methods, \method is constrained only by a global node budget $N$ and constructs the tree iteratively until this capacity is filled. The core of Algorithm \ref{alg:talon} is the Hybrid Expansion Strategy, which dynamically selects the expansion logic based on the current depth. For the root layer ($d=0$), \method employs a Robust Tree Initialization via a standard Top-K expansion to mitigate potential mis-calibration in the draft model's initial prediction. For all subsequent layers ($d \ge 1$), the algorithm switches to a Confidence-Gated Expansion mechanism. It calculates an anchor confidence $m_d$ within the candidate pool and filters nodes using a relative threshold $\mu$ (i.e., $p(u) \ge \mu \cdot m_d$). This "prune-while-expanding" approach allows the tree topology to adapt between "deep-and-narrow" for deterministic contexts and "shallow-and-wide" for uncertain ones.

\section{Token-Tree Verification}
\label{app:verification}

The verification phase of our framework adheres to the standard paradigm of tree-based speculative decoding, widely adopted in the field \citep{eagle, specinfer}. By utilizing the Tree Attention mechanism, the target LLM verifies the entire draft tree in a single forward pass. A structured attention mask ensures that each token within the tree attends only to its predecessors along the root-to-leaf path, effectively simulating independent parallel verifications for all candidate branches while maintaining causal consistency.

Following the methodology established in \citet{eagle}, the verification proceeds in three key steps: (1) calculating the posterior probability of each token in the tree given the target model's output; (2) selecting the optimal valid prefix that satisfies the acceptance criteria; and (3) resampling a correction token from the residual distribution at the point of divergence. This rigorous process guarantees that the final output distribution is mathematically identical to that of the target model. We refer readers to the original \eagle paper \citep{eagle} for detailed algorithms and proofs.

\newpage

\section{Additional Motivated Experiments}
\label{app:motivated_exp}

\begin{figure*}[ht]
    \centering
    \begin{subfigure}{0.48\textwidth}
        \centering
        \includegraphics[width=\linewidth]{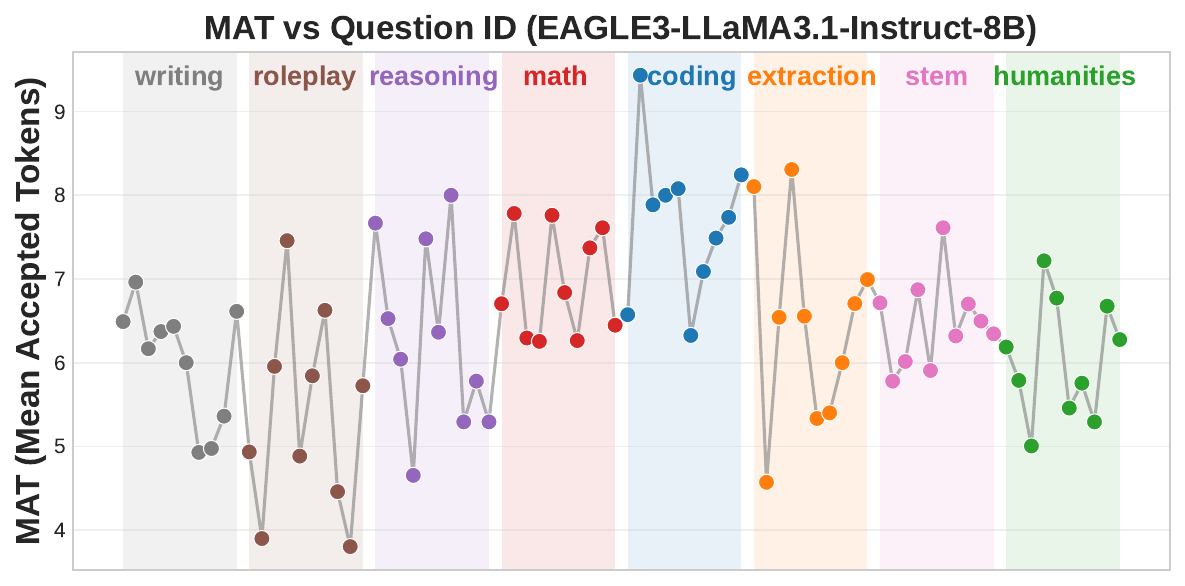}
        \caption{Llama-3.1-Instruct-8B}
        \label{fig:llama3-8b-mat}
    \end{subfigure}
    \hfill
    \begin{subfigure}{0.48\textwidth}
        \centering
        \includegraphics[width=\linewidth]{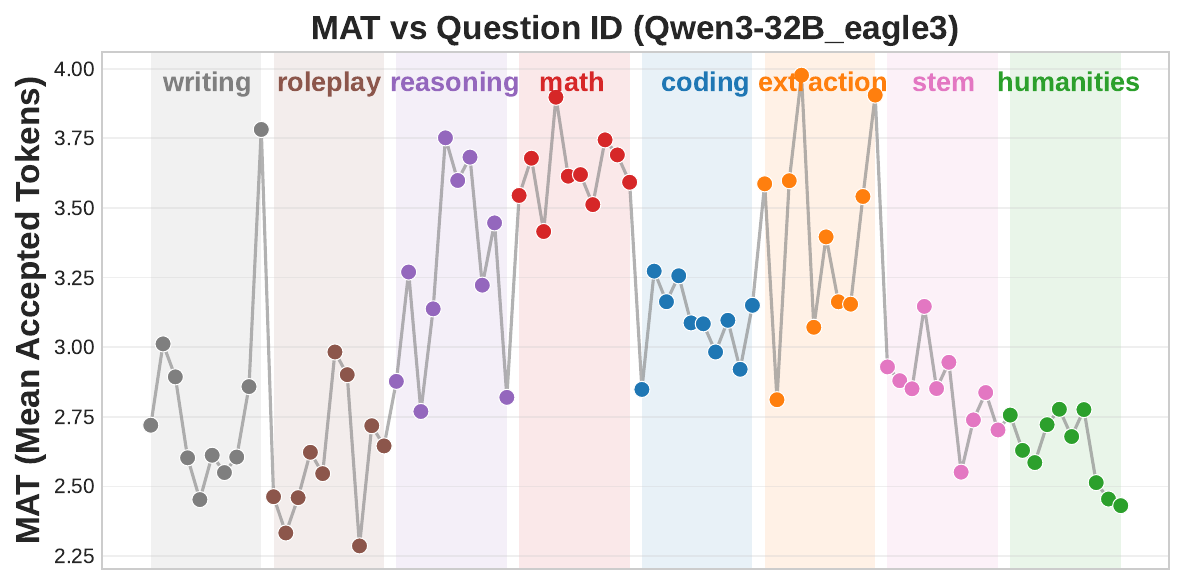}
        \caption{Qwen3-32B}
        \label{fig:qwen3-32b-mat}
    \end{subfigure}
    
    \vspace{1em} 
    
    \begin{subfigure}{0.48\textwidth}
        \centering
        \includegraphics[width=\linewidth]{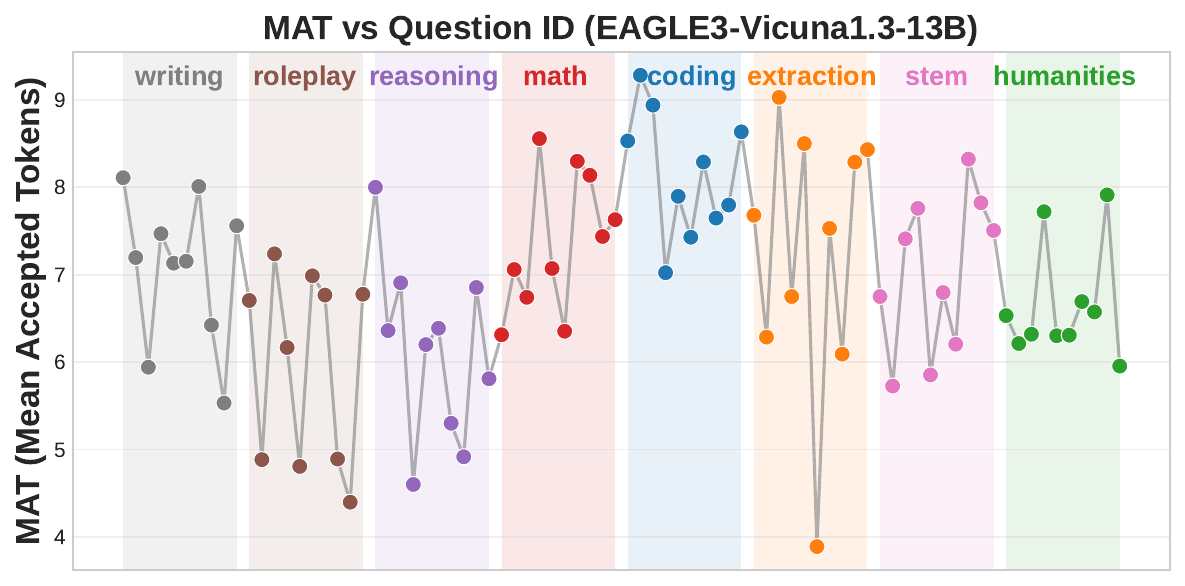}
        \caption{Vicuna-13B}
        \label{fig:vicuna-13b-mat}
    \end{subfigure}
    \hfill
    \begin{subfigure}{0.48\textwidth}
        \centering
        \includegraphics[width=\linewidth]{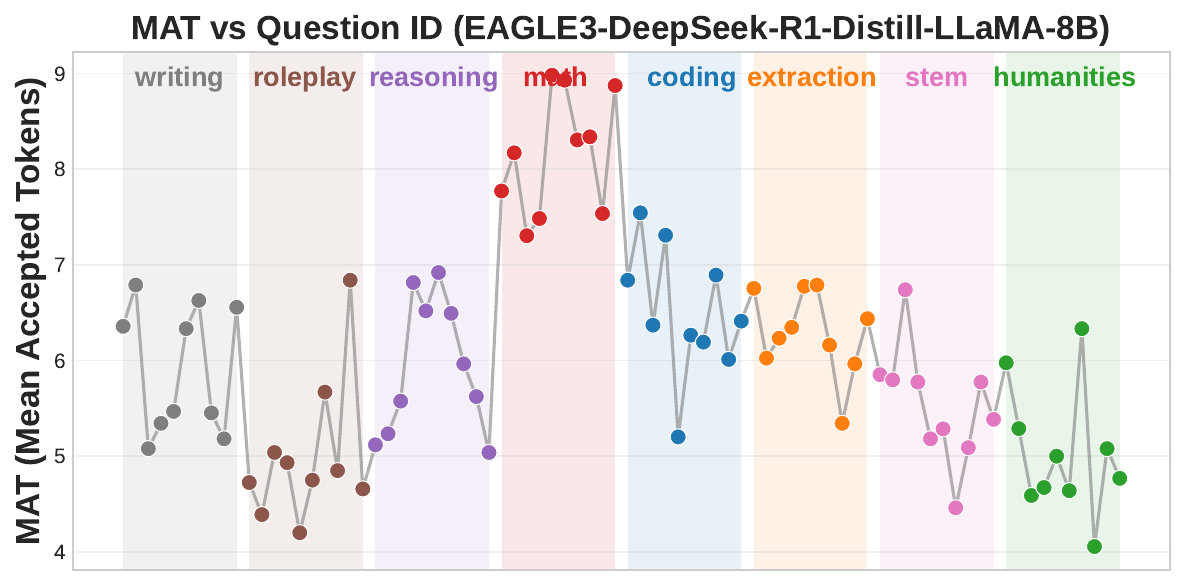}
        \caption{DeepSeek-R1-Distill-LLaMA-8B}
        \label{fig:dsl-8b-mat}
    \end{subfigure}
    
    \caption{Real-World Mean Accepted Tokens (MAT) distribution across different queries with EAGLE baseline. We visualize the MAT fluctuations on (a) Llama-3.1-8B, (b) Qwen3-32B, (c) Vicuna-13B, and (d) DeepSeek-R1-Distill-8B. Across varying parameter scales and architectures, we consistently observe significant volatility in generation difficulty: low-entropy tasks (e.g., Math/Coding) often allow for high acceptance, while high-entropy tasks (e.g., Roleplay) necessitate shallow trees. This universal variance highlights the limitation of static fixed-depth policies and underscores the necessity of \method's adaptive strategy.}
    \label{fig:mat_distribution_all}
\end{figure*}

To demonstrate the generality of the limitations observed in Section 3, specifically the "Acceptance Funnel" phenomenon and the "Static Depth Dilemma," we conduct additional empirical analyses on a broader range of LLMs. These experiments confirm that the inefficiencies of static tree structures are not model-specific artifacts but inherent challenges in speculative decoding.

\subsection{Heat Map Visualization of LlaMA-3.1-Instruct-8B}
\label{app:llama-heat-sec}

\begin{figure}[h]
    \centering
    \includegraphics[width=\linewidth]{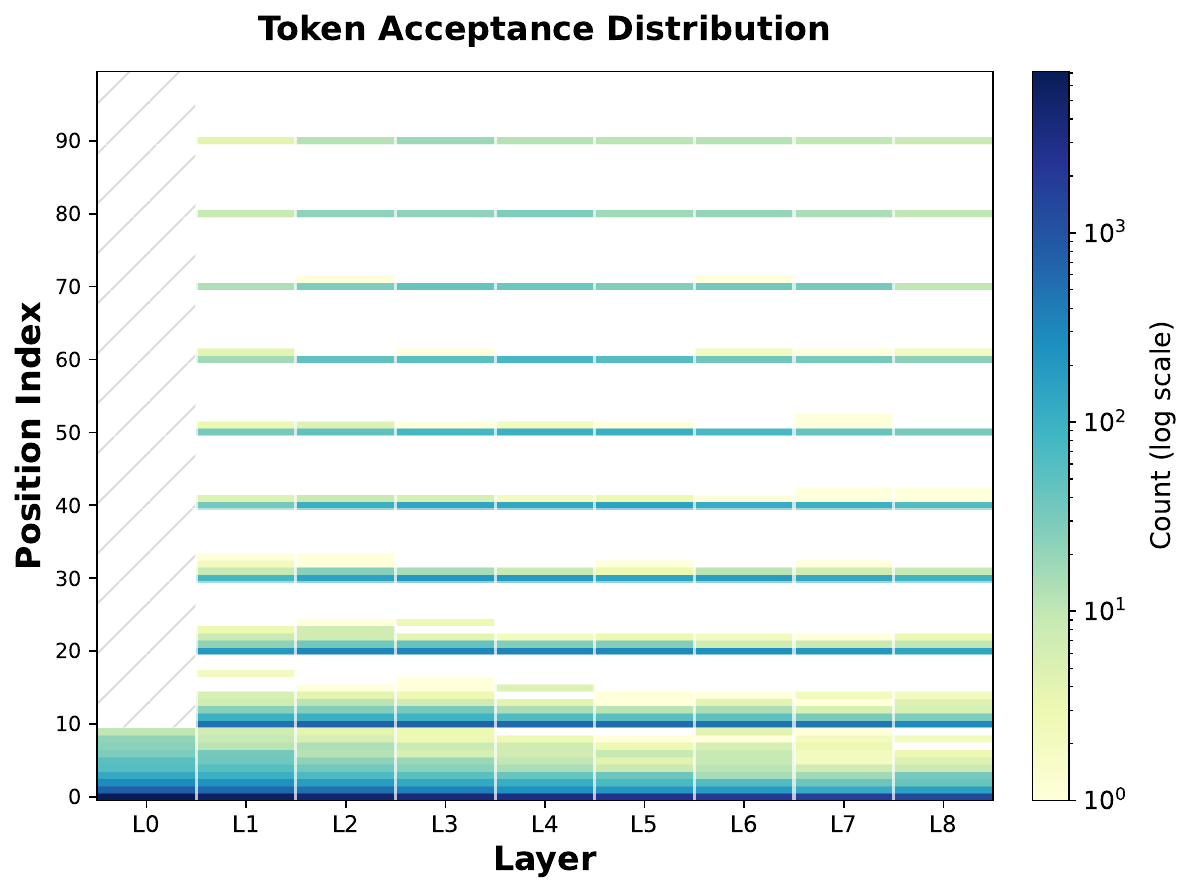}
    \caption{Visualization of token acceptance frequency with Llama-3-8B using a static draft tree ($K=10, D=8$). Consistent with Figure 3, the heatmap demonstrates the ``Acceptance Funnel'' phenomenon: acceptance is dispersed in the first layer but rapidly concentrates on top-ranked candidates in deeper layers, rendering wide static expansion inefficient.}
    \label{fig:llama-heat}
\end{figure}

In Section~\ref{sec:motivation}, we utilized Qwen3-8B to illustrate the funnel-like distribution of accepted tokens. To verify whether this pattern holds across different architectures, we perform the same visualization on Llama-3.1-8B-Instruct. As shown in Figure~\ref{fig:llama-heat}, the results exhibit a striking similarity to our previous findings. The acceptance probability in the initial layer is relatively dispersed, necessitating a robust search width to capture correct continuations. However, as the tree deepens, the acceptance mass concentrates sharply on the high-confidence regions. This consistent "Acceptance Funnel" across models further justifies \method's hybrid expansion strategy: enforcing robustness at the root while employing confidence-gated pruning at deeper layers to eliminate the redundancy of static width.

\subsection{Dynamic Accepted Length of Various Models}
\label{app:mat-sec}

We further extend the analysis of the "Static Depth Dilemma" to evaluate how generation difficulty fluctuates across different models and tasks. Figure \ref{fig:mat_distribution_all} presents the Mean Accepted Tokens (MAT) distribution for Llama-3.1-8B-Instruct, Qwen3-32B-Instruct, Vicuna-13B, and DeepSeek-R1-Distill-LLaMA-8B, respectively.

Across all evaluated models and diverse task categories, we observe significant volatility in the effective speculation length. In deterministic, logic-constrained tasks such as \textbf{Math and Coding}, the MAT frequently reaches high peaks. In these low-entropy scenarios, the draft model is often confident and accurate, yet a pre-defined static depth limit artificially caps the potential speculation length, preventing the system from fully exploiting the easy context for maximum speedup. Conversely, in open-ended or high-entropy tasks like \textbf{Roleplay or Writing}, the acceptance length often drops significantly due to higher uncertainty. Here, a rigid static tree forces the draft model to hallucinate deep branches that are destined for rejection, resulting in wasted computation. This universal variance underscores that a "one-size-fits-all" static tree structure is fundamentally suboptimal, highlighting the necessity of \method's budget-driven adaptive mechanism that dynamically adjusts tree depth based on real-time confidence.

\section{Derivation of Draft Efficiency}
\label{app:derivation}

In this section, we provide the detailed derivation of the wall-time speedup formula presented in Section~\ref{sec:theoretical-analysis}. We define the wall-clock time for a single forward pass of the draft model and the target model as $T_q$ and $T_p$, respectively. Let $L$ denote the total number of tokens in the final generated output sequence. Throughout the generation process, the draft model executes a total of $N_q$ forward passes, while the target model executes $N_p$ verification steps.

In standard Auto-Regressive (AR) decoding, the target model generates tokens sequentially, resulting in a total inference latency:
\begin{equation}
    T_{AR} = L \cdot T_p,
\end{equation}
In contrast, Speculative Decoding (SD) decouples the process into drafting and verification phases. The total latency is the sum of time spent on both phases, expressed as:
\begin{equation}
    T_{SD} = N_p \cdot T_p + N_q \cdot T_q.
\end{equation}
The wall-time speedup $R$ is strictly defined as the ratio of the baseline latency to the speculative latency:
\begin{equation}
    R = \frac{T_{AR}}{T_{SD}} = \frac{L \cdot T_p}{N_p \cdot T_p + N_q \cdot T_q}. \label{eq:speedup}
\end{equation}
To relate this speedup to algorithmic efficiency, we simplify the fraction by dividing both the numerator and the denominator by $N_p \cdot T_p$. We introduce three key metrics: (1) the relative model cost $c$: 
\begin{equation}
    c = \frac{T_q}{T_p},    
\end{equation}
(2) the Mean Accepted Tokens $\tau$:
\begin{equation}
    \tau = \frac{L}{N_p},
\end{equation}
and (3) the Draft Efficiency $\delta$
\begin{equation}
     \delta = \frac{N_q}{N_p},
\end{equation}
which quantifies the computation invested in drafting per verification step. Substituting these variables into Equation~\ref{eq:speedup} yields:
\begin{equation}
    R = \frac{\frac{L}{N_p}}{1 + \frac{N_q}{N_p} \cdot \frac{T_q}{T_p}} = \frac{\tau}{1 + c \cdot \delta}.
\end{equation}
This relationship highlights the advantage of \method. Static methods enforce a fixed depth $D$, locking draft efficiency to a constant $\delta = D + 1$. Consequently, when the acceptance length $\tau$ drops in difficult contexts, the speedup $R$ degrades significantly. In contrast, \method dynamically adjusts the tree size, reducing $\delta$ in uncertain scenarios. By ensuring the draft cost $\delta$ scales down alongside $\tau$, \method preserves a robust speedup $R$ across varying generation difficulties.

\section{Details of Experiments}

\subsection{Implementation Details}
\label{app:implementation}

\subsubsection{Comparison Baselines}
\label{app:baselines}

To strictly evaluate the effectiveness of \method, we compare it against a comprehensive set of competitive baselines, categorizing them into chain-based, tree-based and other MLP-based speculative decoding approaches. We first include standard \textbf{Speculative Decoding (SD)}~\citep{sps1,sps2} as the fundamental chain-based baseline to measure the raw speedup gain over auto-regressive decoding. For MLP-based methods, which aim to reuse target hidden states with MLP to predict multiple draft tokens, we compare against \textsc{Medusa}~\citep{medusa} and \textsc{Hydra}~\citep{hydra}. These methods employ lightweight MLP decoding heads to predict subsequent tokens in parallel, representing a distinct direction in efficiently drafting.

\begin{table*}[t]
    \centering
    \small
    \resizebox{\linewidth}{!}{
    \begin{tabular}{l l l}
    \toprule
    \textbf{Target Model} & \textbf{Method} & \textbf{Draft Model Checkpoint (Hugging Face)} \\
    \midrule
    \multirow{4}{*}{Vicuna-1.3-13B} 
        & Medusa & \texttt{FasterDecoding/medusa-vicuna-13b-v1.3} \\
        & Hydra & \texttt{ankner/hydra-vicuna-13b-v1.3} \\
        & SD (Standard) & \texttt{double7/vicuna-68m} \\
        & OPT-Tree / \eagle-3 / \method & \texttt{yuhuili/EAGLE3-Vicuna1.3-13B} \\
    \midrule
    DSL-8B & \eagle-3 / \method & \texttt{yuhuili/EAGLE3-DeepSeek-R1-Distill-LLaMA-8B} \\
    \midrule
    Llama-3.1-8B-Instruct & \eagle-3 / \method & \texttt{yuhuili/EAGLE3-LLaMA3.1-Instruct-8B} \\
    \midrule
    Qwen3-8B & \eagle-3 / \method & \texttt{AngelSlim/Qwen3-8B\_eagle3} \\
    \midrule
    Qwen3-32B & \eagle-3 / \method & \texttt{AngelSlim/Qwen3-32B\_eagle3} \\
    \bottomrule
    \end{tabular}
    }
    \caption{
        List of draft model checkpoints used in our experiments. 
        \method shares the identical draft weights with \eagle-3 across all tested benchmarks to ensure a fair evaluation of the tree topology efficiency.
    }
    \label{tab:draft_models}
\end{table*}

In the realm of tree-based speculative decoding, we select \eagle-3~\citep{eagle3} as the primary state-of-the-art baseline. \eagle-3 utilizes feature-level auto-regression with a multi-layer fusion mechanism and typically constructs a static draft tree with fixed width and depth. Comparing against \eagle-3 allows us to directly demonstrate the advantages of our adaptive topology over rigid geometric constraints.

Most importantly, we include \textbf{OPT-Tree}~\citep{opt-tree} as a key baseline, as it represents the related work most closely aligned with \method. Similar to our approach, OPT-Tree aims to optimize the draft tree topology. However, a critical distinction lies in the construction paradigm: OPT-Tree typically relies on search-based heuristics or a ``generate-then-prune'' strategy, which can introduce non-trivial computational overhead during inference. In contrast, \method adopts a training-free, budget-driven expansion strategy that dynamically shapes the tree on the fly based on real-time confidence, thereby achieving adaptivity without the latency costs associated with complex search algorithms.

\begin{figure*}[ht]
    \centering
    \begin{subfigure}{0.48\textwidth}
        \centering
        \includegraphics[width=\linewidth]{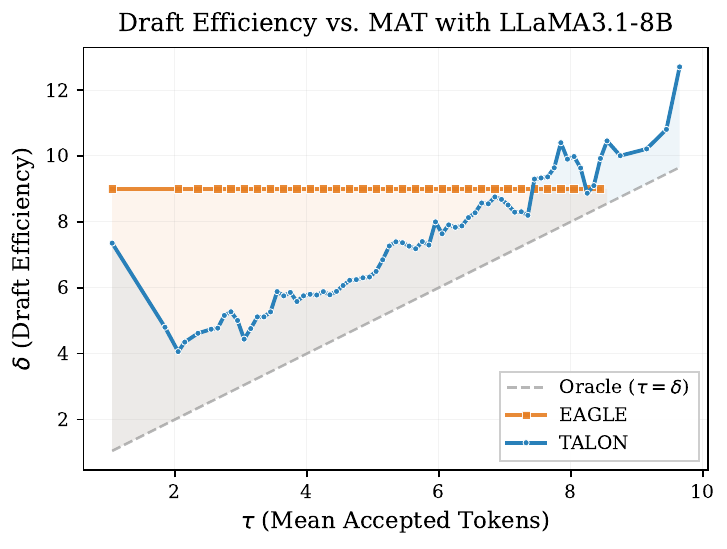}
        \caption{Llama-3.1-Instruct-8B}
        \label{fig:draft-efficiency-llama3-8b}
    \end{subfigure}
    \hfill
    \begin{subfigure}{0.48\textwidth}
        \centering
        \includegraphics[width=\linewidth]{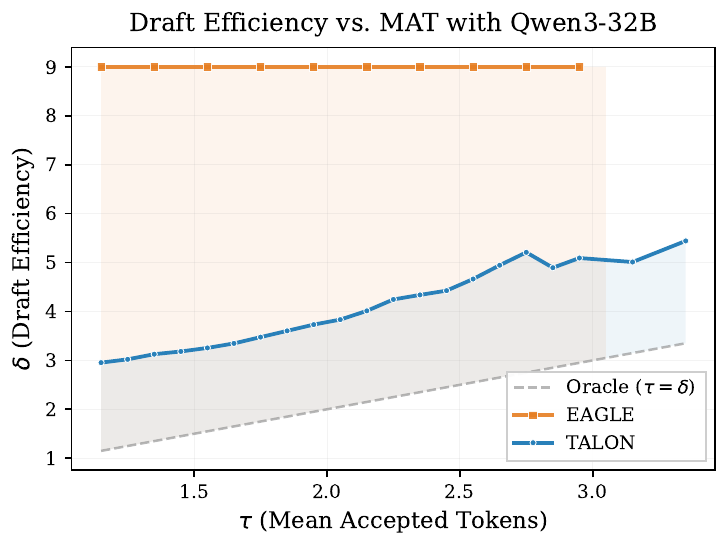}
        \caption{Qwen3-32B}
        \label{fig:draft-efficiency-qwen3-32b}
    \end{subfigure}
    
    \vspace{1em} 
    
    \begin{subfigure}{0.48\textwidth}
        \centering
        \includegraphics[width=\linewidth]{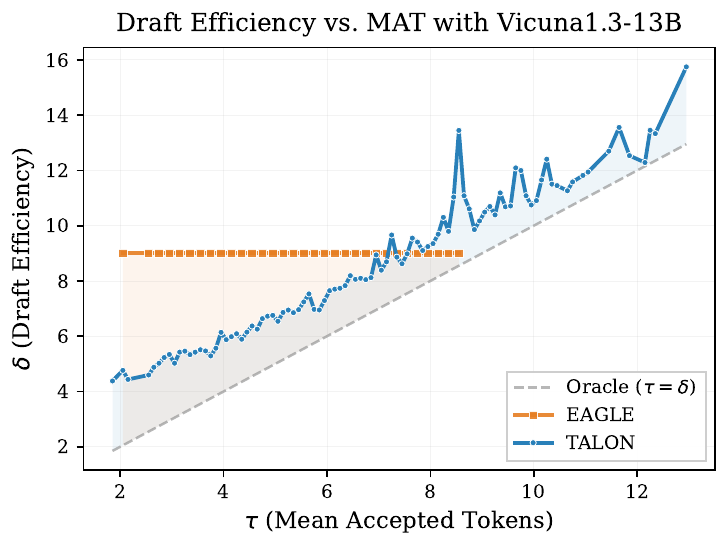}
        \caption{Vicuna-13B}
        \label{fig:draft-efficiency-vicuna-13b}
    \end{subfigure}
    \hfill
    \begin{subfigure}{0.48\textwidth}
        \centering
        \includegraphics[width=\linewidth]{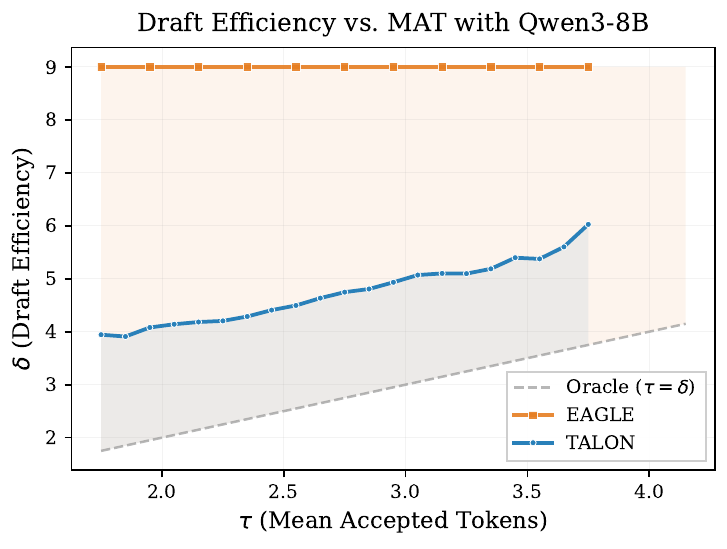}
        \caption{Qwen3-8B}
        \label{fig:draft-efficiency-qwen3-8b}
    \end{subfigure}
    
    \caption{\textbf{Detailed visualization of Draft Efficiency ($\delta$) versus Mean Accepted Tokens ($\tau$) across four different LLMs.} The x-axis represents the mean accepted length, while the y-axis represents the computational cost (draft steps per verification). The gray dashed line denotes the optimal Oracle baseline ($\tau = \delta$) where no computation is wasted. While the static \eagle-3 baseline (orange) maintains a fixed high cost regardless of difficulty, TALON (blue) dynamically adjusts its draft budget, closely tracking the Oracle curve and significantly reducing computational waste in high-entropy (low $\tau$) scenarios and increasing acceptance reward in low-entropy (high $\tau$) scenarios.}
    \label{fig:draft_efficiency_all}
\end{figure*}

\subsubsection{Hardware and Software Configurations}
\label{app:hardware_setup}

\paragraph{Hardware Environment.} All experiments are conducted on a server equipped with a single \textbf{NVIDIA H200 (141GB) GPU}~\citep{cuda}. We perform all evaluations with a batch size of 1 to follow the standard settings and simulate real-world latency-critical inference scenarios. 

\paragraph{Software Environment.} Our implementation is based on \texttt{PyTorch}~\citep{pytorch} version \texttt{2.6.0+cu124} and \texttt{Transformers}~\citep{transformers} version \texttt{4.57.1}. The code is compiled with CUDA \texttt{12.8}. For SD (Vanilla Speculative Decoding), we use Transformers' official assisted decoding with their default settings. For \textsc{Medusa} and \textsc{Hydra}, we use their official codebases and adhere to their recommended environment settings to ensure fair comparison. For OPT-Tree, as its official codebase only supports \eagle-2, we re-implement OPT-Tree to use official \eagle-3 draft model checkpoint.

\paragraph{Generation Configuration.} Unless otherwise specified, we employ greedy decoding (Temperature $T=0$) for the main speedup benchmarks reported in Table~\ref{tab:main_results}. For the robustness experiments under stochastic sampling (Table~\ref{tab:temp_results}), we set the temperature to $T=1.0$ (No additional top-k or top-p operation). The maximum generation length is set to \texttt{1024} tokens for standard benchmarks. Regarding \method's hyper-parameters, we set the global token budget $N=60$ and the confidence threshold $\mu=0.03$ by default across all models, $K=10$ for robust tree initialization. For \eagle-3, we set the tree width $K=10$ and $D=8$, which is reported as optimal values in their manuscripts.

\begin{figure*}[htbp]
    \centering
    \includegraphics[width=\textwidth]{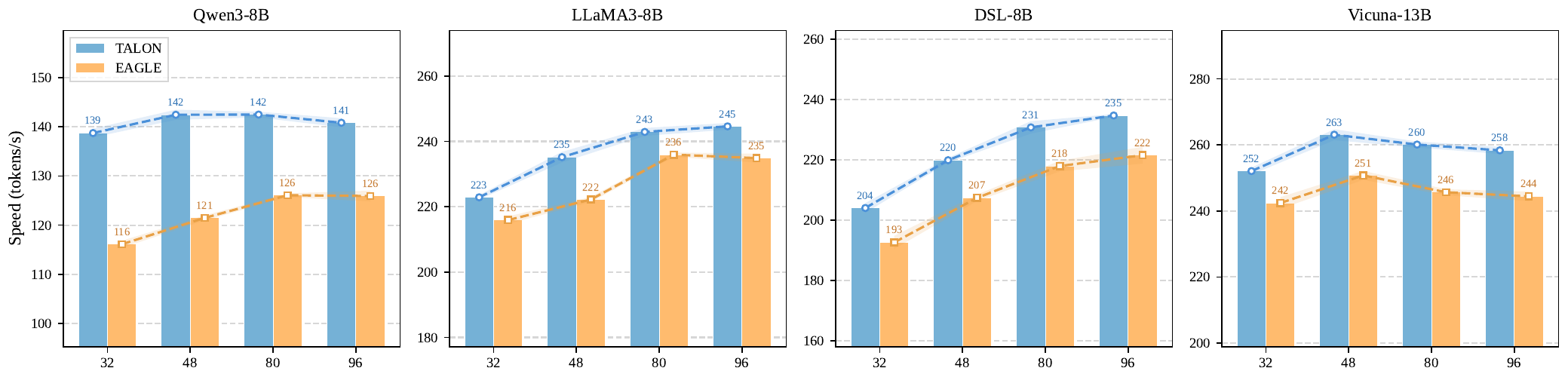}
    \caption{\textbf{Ablation study on the global token budget ($N$).} We compare the wall-time generation speed of \method against the \eagle-3 baseline across varying budget constraints ranging from $N=32$ to $N=96$. The results demonstrate that \method consistently outperforms the static baseline across all budget levels. The performance gap is particularly pronounced in resource-constrained settings (e.g., $N=32$), confirming that \method's confidence-gated expansion strategy is significantly more efficient at prioritizing high-value draft tokens than the rigid "expand-then-shrink" mechanism of static methods.}
    \label{fig:ablation-total-token}
\end{figure*}

\subsubsection{Draft Model Selection}
\label{app:draft_model_selection}

To ensure full reproducibility and a strictly fair comparison, we list the exact Hugging Face model checkpoints used for all draft models in our experiments. 
Crucially, for \textsc{Talon}, we utilize the \textit{exact same} draft model checkpoints as the state-of-the-art baseline EAGLE-3 (and OPT-Tree where applicable). This ensures that any observed performance gains are attributed solely to our adaptive tree expansion algorithm rather than superior draft weights.
The detailed configurations are provided in Table~\ref{tab:draft_models}.

\subsection{More Visualization Results of Draft Efficiency}
\label{app:draft_efficiency}

To comprehensively validate the theoretical efficiency analysis presented in Section~\ref{sec:theoretical-analysis}, we provide extended visualizations of the relationship between Draft Efficiency ($\delta$) and Mean Accepted Tokens ($\tau$) across varying model architectures. Figure~\ref{fig:draft_efficiency_all} illustrates this correlation for Llama-3.1-8B-Instruct, Qwen3-32B, Vicuna-13B, and Qwen3-8B. In these plots, the x-axis represents the acceptance length (how many tokens are actually accepted within each decoding step), while the y-axis represents the computational overhead draft efficiency (the ratio of draft steps to verification steps). The gray dashed line serves as the Oracle baseline ($\tau = \delta$), representing an ideal zero-waste scenario where every drafted token is accepted.

As clearly demonstrated across all four sub-figures, the static baseline (\eagle-3) exhibits a rigid, horizontal trajectory. This pattern confirms that static tree-based methods incur a constant computational overhead determined solely by their fixed geometric hyperparameters (width $K$ and depth $D$), regardless of the actual generation difficulty. Consequently, in high-entropy scenarios where the acceptance length $\tau$ is low, static methods suffer from a substantial "efficiency gap"—highlighted by the extensive orange shaded regions—indicating that the system is squandering computational resources on generating draft branches that are destined to be rejected.

In sharp contrast, TALON demonstrates a highly adaptive behavior, with its efficiency curve strictly tracking the Oracle baseline across the entire spectrum of generation difficulties. The upward-sloping blue trajectory indicates that TALON dynamically modulates its resource investment: it autonomously reduces the draft budget in uncertain contexts to minimize waste, while scaling up the tree size in deterministic contexts to maximize the acceptance length. This tight alignment between the draft cost ($\delta$) and the acceptance reward ($\tau$) empirically verifies that our confidence-gated expansion strategy effectively decouples the draft structure from rigid constraints, ensuring that computational resources are allocated only when they yield a positive marginal utility. (Notably, the curve of Qwen3-8B and Qwen3-32B is relatively far from oracle line, suggesting that there is still a room to increase \method's end-to-end speedup. We will show in following sections that decreasing \method's threshold from default \texttt{0.03} to smaller threshold \texttt{0.01} yields better performance.)

\begin{figure*}[htbp]
    \centering
    \includegraphics[width=\textwidth]{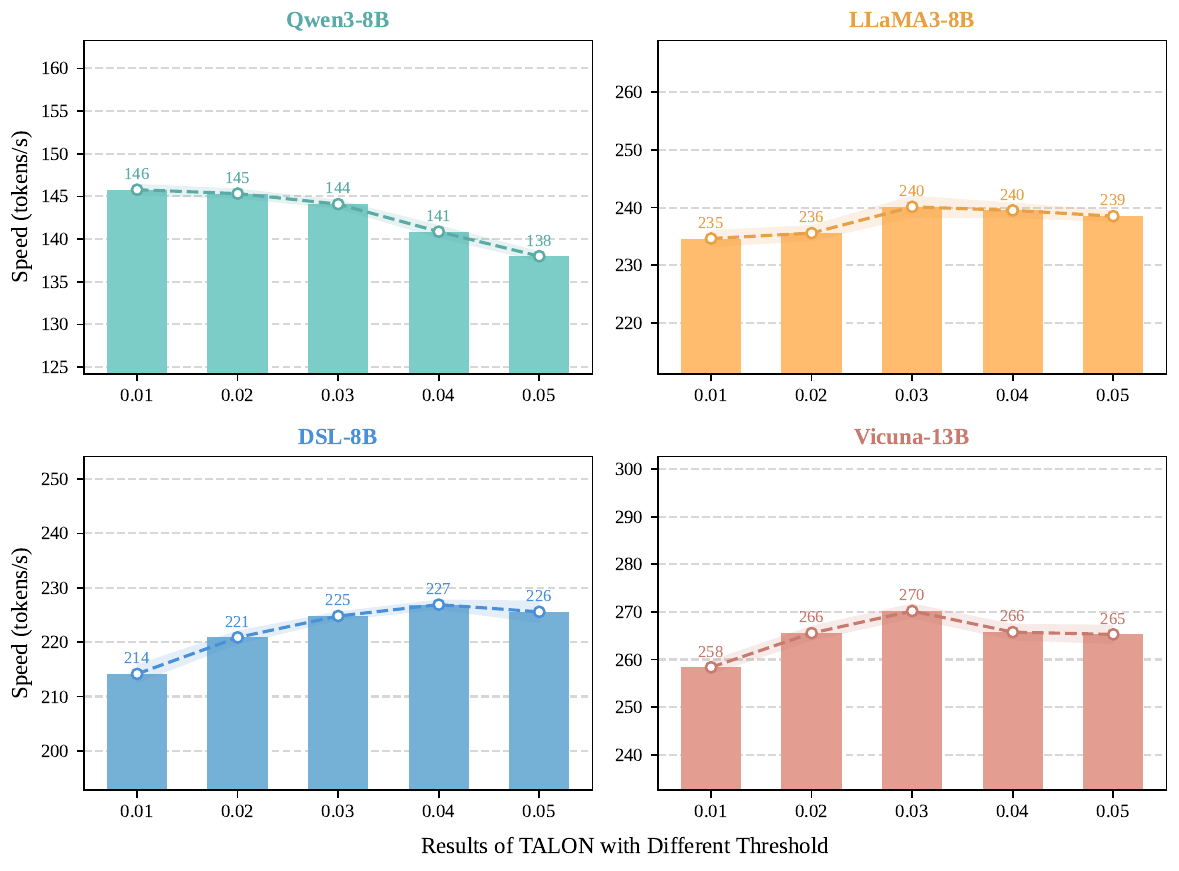}
    \caption{\textbf{Sensitivity analysis of the confidence threshold $\mu$.} This parameter controls the trade-off between exploration width and generation depth. The results indicate that the optimal $\mu$ is positively correlated with the model's draft-target alignment (MAT). For models with lower alignment like Qwen3-8B (a), a lower threshold ($\mu=0.01$) yields the best performance by encouraging a "shallow-and-wide" search to ensure coverage. In contrast, for highly aligned models like DSL-8B (c), a higher threshold ($\mu=0.04$) is preferred to form "deep-and-narrow" chains that maximize the speculation length.}
    \label{fig:ablation-threshold}
\end{figure*}

\begin{table*}[t]
\centering
\renewcommand{\arraystretch}{1.1}
\caption{\textbf{Main Results on Six Benchmarks (Temperature=1).} Comparison between \eagle-3 and our proposed \textbf{\method} across various models. We report Mean Acceptance Tokens (MAT) and Wall-time Speedup relative to standard decoding. \textbf{Bold} numbers denote the best speedup performance.}
\label{tab:temp_results}

\resizebox{\textwidth}{!}{%
\setlength{\tabcolsep}{4pt}
\begin{tabular}{llcccccccccccc}
\toprule
\multirow{2}{*}{\textbf{Model}} & \multirow{2}{*}{\textbf{Method}} & \multicolumn{2}{c}{\textbf{Alpaca}} & \multicolumn{2}{c}{\textbf{GSM8K}} & \multicolumn{2}{c}{\textbf{HumanEval}} & \multicolumn{2}{c}{\textbf{MT-Bench}} & \multicolumn{2}{c}{\textbf{QA}} & \multicolumn{2}{c}{\textbf{CNN/DM}} \\
\cmidrule(lr){3-4} \cmidrule(lr){5-6} \cmidrule(lr){7-8} \cmidrule(lr){9-10} \cmidrule(lr){11-12} \cmidrule(lr){13-14}
& & \textsc{Mat} & Spd. & \textsc{Mat} & Spd. & \textsc{Mat} & Spd. & \textsc{Mat} & Spd. & \textsc{Mat} & Spd. & \textsc{Mat} & Spd. \\

\midrule

\multirow{6}{*}{Vicuna-13B} & \eagle-3 & 5.61 & 3.15$\times$ & 5.95 & 3.29$\times$ & 6.77 & 3.78$\times$ & 5.79 & 3.21$\times$ & 4.72 & 2.64$\times$ & 6.01 & 2.96$\times$ \\
 & SD & 1.84 & 1.11$\times$ & 1.84 & 1.08$\times$ & 2.27 & 1.29$\times$ & 2.04 & 1.13$\times$ & 1.71 & 1.05$\times$ & 2.21 & 1.23$\times$ \\
 & \textsc{Medusa} & 2.83 & 2.19$\times$ & 2.82 & 2.20$\times$ & 2.97 & 2.35$\times$ & 2.84 & 2.22$\times$ & 2.49 & 1.92$\times$ & 2.26 & 1.69$\times$ \\
 & \textsc{Hydra} & 4.23 & 2.86$\times$ & 3.98 & 2.73$\times$ & 4.18 & 2.91$\times$ & 4.06 & 2.77$\times$ & 3.31 & 2.25$\times$ & 3.15 & 2.07$\times$ \\
 & OPT-Tree & 5.56 & 3.21$\times$ & 5.97 & 3.25$\times$ & 6.47 & 3.53$\times$ & 5.69 & 3.16$\times$ & 4.86 & 2.77$\times$ & 5.84 & 2.79$\times$ \\
 & \green \textbf{\method} & \green 5.43 & \green \textbf{3.28$\times$} & \green 5.82 & \green \textbf{3.35$\times$} & \green 7.25 & \green \textbf{3.97$\times$} & \green 5.88 & \green \textbf{3.38$\times$} & \green 4.63 & \green \textbf{2.82$\times$} & \green 5.95 & \green \textbf{2.96$\times$} \\

\midrule
\multirow{2}{*}{DSL-8B} & \eagle-3 & 4.48 & 2.49$\times$ & 6.43 & 3.58$\times$ & 5.47 & 3.04$\times$ & 4.56 & 2.55$\times$ & 4.04 & 2.25$\times$ & 4.32 & 2.39$\times$ \\
 & \green \textbf{\method} & \green 4.29 & \green \textbf{2.75$\times$} & \green 6.44 & \green \textbf{3.70$\times$} & \green 5.30 & \green \textbf{3.25$\times$} & \green 4.44 & \green \textbf{2.81$\times$} & \green 3.91 & \green \textbf{2.54$\times$} & \green 4.17 & \green \textbf{2.63$\times$} \\

\midrule

\multirow{2}{*}{Llama3-8B} & \eagle-3 & 5.19 & 2.85$\times$ & 4.59 & 2.51$\times$ & 6.24 & 3.46$\times$ & 3.96 & 2.20$\times$ & 3.12 & 1.72$\times$ & 4.44 & 2.42$\times$ \\
 & \green \textbf{\method} & \green 5.21 & \green \textbf{3.15$\times$} & \green 4.74 & \green \textbf{2.84$\times$} & \green 6.40 & \green \textbf{3.56$\times$} & \green 4.13 & \green \textbf{2.58$\times$} & \green 3.20 & \green \textbf{2.08$\times$} & \green 4.33 & \green \textbf{2.62$\times$} \\

\midrule

\multirow{2}{*}{Qwen3-8B} & \eagle-3 & 3.32 & 1.95$\times$ & 3.81 & 2.24$\times$ & 3.77 & 2.20$\times$ & 3.46 & 2.02$\times$ & 3.17 & 1.87$\times$ & 3.17 & 1.85$\times$ \\
 & \green \textbf{\method} & \green 3.28 & \green \textbf{2.25$\times$} & \green 3.74 & \green \textbf{2.50$\times$} & \green 3.70 & \green \textbf{2.51$\times$} & \green 3.41 & \green \textbf{2.32$\times$} & \green 3.15 & \green \textbf{2.17$\times$} & \green 3.11 & \green \textbf{2.11$\times$} \\

\midrule

\multirow{2}{*}{Qwen3-32B} & \eagle-3 & 2.55 & 1.67$\times$ & 3.22 & 2.10$\times$ & 2.95 & 1.89$\times$ & 2.74 & 1.75$\times$ & 2.44 & 1.61$\times$ & 2.46 & 1.51$\times$ \\
 & \green \textbf{\method} & \green 2.53 & \green \textbf{1.84$\times$} & \green 3.17 & \green \textbf{2.27$\times$} & \green 2.91 & \green \textbf{2.07$\times$} & \green 2.71 & \green \textbf{1.92$\times$} & \green 2.41 & \green \textbf{1.77$\times$} & \green 2.43 & \green \textbf{1.64$\times$} \\

\bottomrule
\end{tabular}%
}
\end{table*}

\subsection{More Evaluation Results under Temperature Settings}
\label{app:temp_exp}

While the primary evaluations in the main text focus on greedy decoding (Temperature $T=0$), real-world LLM applications—particularly creative writing and open-ended chat—frequently rely on stochastic sampling to induce diversity in the generated content. To rigorously assess the robustness of our framework in these non-deterministic scenarios, we conduct a comprehensive evaluation using standard sampling with Temperature $T=1$. The detailed comparative results between \method and the state-of-the-art baseline \eagle-3 are reported in Table~\ref{tab:temp_results}.

As evidenced by the quantitative results, \method consistently outperforms \eagle-3 across all evaluated models and benchmarks, demonstrating superior adaptability to stochastic environments. A fundamental challenge in high-temperature settings is that the target model's probability distribution becomes flatter, reducing the dominance of the top-1 token and increasing the likelihood of selecting lower-ranked candidates. Static approaches like \eagle-3, which enforce a fixed width and depth, often struggle in this regime because they rigidly expand the top-$K$ candidates regardless of the entropy. This leads to a scenario where the draft tree either misses the sampled token due to insufficient coverage in uncertain branches or wastes computation on high-confidence paths that are not selected.

In contrast, \method's confidence-gated expansion strategy naturally excels under these conditions. By determining the tree topology based on relative probability thresholds rather than a fixed node count, TALON dynamically widens the search space in high-entropy layers to capture the dispersed probability mass, while keeping the tree narrow in clearer contexts. This flexibility is reflected in the substantial speedup gains; for instance, on the Vicuna-13B model for HumanEval, \method achieves a wall-time speedup of 3.97$\times$, significantly surpassing the 3.53$\times$ speedup of \eagle-3. Similarly, on Qwen3-8B, \method maintains a distinct advantage in both reasoning-intensive tasks like GSM8K and open-ended tasks like Alpaca. These findings confirm that \method's budget-driven mechanism is not limited to deterministic acceleration but is a robust solution for diverse generation settings, effectively mitigating the efficiency degradation often observed in stochastic speculative decoding.

\subsection{More Ablation Studies of \method}
\label{app:ablation}

In this section, we conduct in-depth ablation studies to evaluate the sensitivity and robustness of \method with respect to its two key hyper-parameters: the global token budget ($N$) and the confidence threshold ($\mu$).

\begin{figure*}[t]
    \centering
    \setlength{\fboxsep}{1.2pt}
    
    \footnotesize
    \begin{tabular}{cc}
        \tokacc{\phantom{x}} Draft token (accepted) & 
        \tokrej{\phantom{x}} Target token (rejected \& resampled)
    \end{tabular}
    \vspace{0.5em}

    \begin{tabular}{|p{0.48\linewidth}|p{0.48\linewidth}|}
        \hline
        \rowcolor{gray!10} 
        \eagle-3 \hfill \footnotesize (Tokens: 197 | Acc: 88.8\%) & 
         \method \hfill \footnotesize (Tokens: 197 | Acc: \textbf{92.4\%}) \\
        \hline
        
        \vspace{2pt}
        \small
        \tokacc{<s>}\tokacc{ Here}\tokacc{'}\tokacc{s}\tokacc{ the}\tokacc{ completed}\tokacc{ code}\tokacc{:}\newline
        \tokacc{}\tokrej{```}\tokacc{python}\newline
        \tokacc{def}\tokacc{ next}\tokacc{\_}\tokacc{small}\tokacc{est}\tokacc{(}\tokacc{lst}\tokrej{):}\newline
        \tokacc{   }\tokacc{ """}\newline
        \tokacc{   }\tokacc{ You}\tokacc{ are}\tokacc{ given}\tokacc{ a}\tokrej{ list}\tokacc{ of}\tokacc{ integers}\tokacc{.}\newline
        \tokacc{   }\tokacc{ Write}\tokacc{ a}\tokacc{ function}\tokacc{ next}\tokrej{\_}\tokrej{small}\tokacc{est}\tokacc{()}\tokacc{ that}\toktxt{...}\newline
        \tokacc{   }\tokacc{ Return}\tokacc{ None}\tokacc{ if}\tokrej{ there}\tokacc{ is}\tokacc{ no}\tokacc{ such}\tokacc{ element}\tokacc{.}\newline
        \toktxt{}\newline
        \tokacc{   }\tokrej{ next}\tokacc{\_}\tokacc{small}\tokacc{est}\tokacc{([}\tokacc{1}\tokacc{,}\tokacc{ }\tokacc{2}\tokacc{,}\tokrej{ 3}\tokacc{,}\tokacc{ 4}\tokacc{,}\tokacc{ 5}\tokacc{])}\tokacc{ ==}\tokrej{ }\tokacc{2}\newline
        \vspace{2pt}
        & 
        
        \vspace{2pt}
        \small
        \tokacc{<s>}\tokacc{ Here}\tokacc{'}\tokacc{s}\tokacc{ the}\tokacc{ completed}\tokrej{ code}\tokacc{:}\newline
        \tokacc{```}\tokacc{python}\newline
        \tokacc{def}\tokacc{ next}\tokacc{\_}\tokacc{small}\tokacc{est}\tokacc{(}\tokacc{lst}\tokacc{):}\newline
        \tokrej{   }\tokacc{ """}\newline
        \tokacc{   }\tokacc{ You}\tokacc{ are}\tokacc{ given}\tokacc{ a}\tokacc{ list}\tokacc{ of}\tokacc{ integers}\tokacc{.}\newline
        \tokacc{   }\tokacc{ Write}\tokrej{ a}\tokacc{ function}\tokacc{ next}\tokacc{\_}\tokacc{small}\tokacc{est}\tokacc{()}\tokacc{ that}\toktxt{...}\newline
        \tokacc{   }\tokacc{ Return}\tokrej{ None}\tokacc{ if}\tokacc{ there}\tokacc{ is}\tokacc{ no}\tokacc{ such}\tokacc{ element}\tokacc{.}\newline
        \toktxt{}\newline
        \tokacc{   }\tokrej{ next}\tokacc{\_}\tokacc{small}\tokacc{est}\tokacc{([}\tokacc{1}\tokacc{,}\tokacc{ }\tokacc{2}\tokacc{,}\tokacc{ 3}\tokacc{,}\tokacc{ 4}\tokacc{,}\tokacc{ 5}\tokacc{])}\tokacc{ ==}\tokrej{ }\tokacc{2}\newline
        \vspace{2pt}
        \\
        \hline
    \end{tabular}
    
    \caption{\label{fig:token_viz}Comparison of token generation traces. \textbf{Left:} \eagle-3 shows frequent interruptions (pink) due to verification failures. \textbf{Right:} \method maintains longer accepted chains (cyan) by utilizing confidence-aware adaptive trees.}
\end{figure*}

\paragraph{Impact of Global Token Budget ($N$).}
We first investigate how the inference performance scales with the available computational budget. Figure~\ref{fig:ablation-total-token} compares the wall-time speedup of \method against the \eagle-3 baseline across varying node budgets ranging from 32 to 96. For a fair comparison, we apply the same global budget constraint to \eagle-3 using its standard pruning mechanism. The results demonstrate that \method consistently outperforms the static baseline across all budget levels. Notably, in resource-constrained scenarios (e.g., $N=32$), TALON exhibits a significant advantage. This indicates that our confidence-gated expansion is highly efficient at prioritizing the most promising candidate tokens, ensuring that a limited budget is invested in high-probability paths rather than being diluted by a fixed-width expansion. As the budget increases to 96, \method continues to scale effectively, utilizing the additional capacity to extend generation depth in deterministic regions, whereas static methods often hit a performance plateau due to their rigid structural constraints.

\paragraph{Sensitivity to Confidence Threshold ($\mu$).}
Next, we analyze the influence of the confidence threshold $\mu$ on generation speed, as visualized in Figure~\ref{fig:ablation-threshold}. The threshold $\mu$ acts as a gatekeeper that balances the trade-off between exploration width and generation depth. Interestingly, we observe that the optimal $\mu$ is positively correlated with the degree of alignment between the draft and target models, which can be approximated by the Mean Accepted Tokens (MAT). 

For models exhibiting high alignment and high MAT, such as DeepSeek-R1-Distill-LLaMA-8B (DSL-8B), the optimal threshold tends to be relatively high (peaking around $\mu=0.04$). In these well-aligned scenarios, the draft model's confidence is a reliable proxy for correctness; thus, a stricter threshold effectively filters out unlikely branches, shaping the tree into a "deep-and-narrow" structure that maximizes draft length. Conversely, when the draft-target alignment is weaker (resulting in lower MAT), a lower threshold becomes more advantageous. This phenomenon explains the behavior of Qwen3-8B in Figure~\ref{fig:ablation-threshold}, where the performance peaks at a lower threshold of $\mu=0.01$. Here, the draft model is less certain, necessitating a more lenient threshold to encourage "shallow-and-wide" exploration, thereby ensuring sufficient coverage of the target distribution to prevent early rejection.

\subsection{More Case Studies of \method}
\label{app:case_study}

To qualitatively analyze the behavior of \method, we visualize the decoding traces of a code generation task in Figure~\ref{fig:token_viz}. Code generation typically represents a low-entropy regime, where the next tokens (e.g., syntax keywords, standard indentations) can be predicted with high confidence.

As shown in the left panel, \eagle employs a relatively static expansion strategy. Even when the model is confident, \eagle is limited by its fixed tree structure, resulting in frequent interruptions (pink tokens) and forcing the model to resample from the target model. This limits the Mean Accepted Tokens (MAT) per step.

In contrast, \method (right panel) dynamically leverages its confidence-gated expansion mechanism. Upon detecting the low-entropy nature of the current context, \method adaptively allocates the token budget to extend the tree \textit{depth} rather than width. This allows the draft model to speculate deep-and-narrow draft tokens. Consequently, \method significantly reduces the frequency of verification calls and achieves a higher MAT, demonstrating the efficiency of adaptive token trees in deterministic generation tasks.

\subsection{Tree Construction Overhead}
\label{app:tree-cons-overhead}

To rigorously quantify the algorithmic efficiency of our proposed method, we conduct a micro-benchmark focusing specifically on the \textbf{tree construction overhead}. We simulate the next-token probability distributions using a Zipfian distribution $P(r) \propto 1/r^\alpha$ \citep{zipf}, covering diverse generation scenarios ranging from high-entropy creative writing tasks ($\alpha=0.7$) and standard natural language ($\alpha=1.35$) to highly deterministic code generation ($\alpha=5.0$). The warmup step is \texttt{20}, and we run each tree expansion strategy \texttt{100} times and report their mean latency. In this controlled environment, we compare the single-layer expansion latency of \eagle against \method. \eagle relies on a rigid dual Top-K mechanism that performs sorting operations twice per layer—once for child selection and again for parent ranking—which scales poorly with large vocabulary sizes. In contrast, \method utilizes a single confidence-gated sampling operation, implemented via efficient element-wise masking and non-zero index retrieval, which does not involving ranking operation.

\begin{figure*}[htbp]
    \centering
    \includegraphics[width=\textwidth]{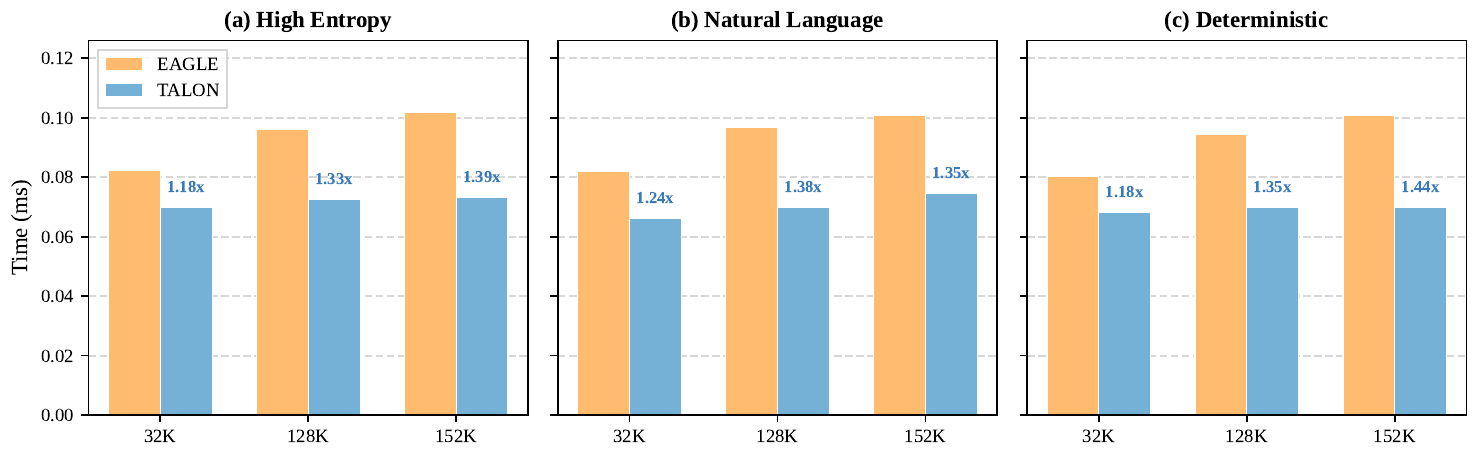}
    \caption{
    \textbf{Runtime breakdown of the tree expansion overhead.} 
        We benchmark the latency of a single-layer expansion step across varying vocabulary sizes ($32K$, $128K$, $152K$) using Zipfian distributions \citep{zipf} to simulate High Entropy ($\alpha=0.7$), Natural Language ($\alpha=1.35$), and Deterministic ($\alpha=5.0$) contexts. 
        \method consistently outperforms the static dual Top-K approach of \eagle, achieving up to \textbf{1.44$\times$} speedup in the construction phase by avoiding expensive sorting operations. \textbf{Note that this step accounts for less than 5\% of the total inference time.}
        }
    \label{fig:tree-overhead}
\end{figure*}

As illustrated in Figure~\ref{fig:tree-overhead}, \method consistently reduces the tree construction latency across varying vocabulary sizes ($32K$ to $152K$), achieving speedups ranging from 1.18$\times$ to 1.44$\times$. The advantage is particularly pronounced in deterministic settings ($\alpha=5.0$), where our adaptive mechanism naturally sparsifies the candidate set, thereby minimizing memory access overhead. However, it is important to acknowledge that \textbf{this sampling and tree construction latency constitutes a negligible fraction ($<5\%$) of the total end-to-end inference time}, which remains dominated by the model's forward passes. Consequently, \method's primary wall-time speedup derives from its superior Draft Efficiency ($\delta$) rather than this micro-optimization in sampling.

\section{More Discussions to Related Work}\label{app:related_work}
This appendix provides an extended discussion on the evolution of speculative decoding, tracing the field's progression from chain-based verification to tree-structured verification. We highlight how varying approaches trade off draft quality, verification cost, and implementation complexity.

\paragraph{Chain-based speculative decoding.}
The classical formulation of speculative decoding, often referred to as speculative sampling, is a \emph{draft-and-verify} procedure \citep{sps1, sps2, draft_and_verify}. Given a prefix, a fast drafter autoregressively proposes a length-$K$ continuation. The target model then verifies these $K$ positions in a single forward pass, accepting the longest matching prefix and resampling at the first mismatch. While attractive for its lossless nature, the speedup of this framework hinges strictly on the acceptance length and the computational cost ratio between the drafter and the target.

\emph{Improving the drafter.}
A major line of work focuses on producing higher-quality drafts with minimal overhead to boost acceptance rates.
Medusa \citep{medusa} introduces multiple lightweight decoding heads atop the target model to predict several next tokens in parallel by reusing target hidden states.
Because independent heads may ignore intra-draft dependencies, follow-up approaches like Hydra \citep{hydra} and Clover \citep{clover} introduce sequential dependencies among heads to better approximate autoregressive drafting.
Chimera \citep{chimera} proposes a lightweight draft architecture combining short-range and full-context signals to enhance quality while maintaining speed.
Complementarily, GliDe with CaPE \citep{glide} reuses the target KV cache to lower drafting overhead and employs confidence-guided proposal expansion to provide stronger candidates.

\emph{Controlling draft length and reducing wasted work.}
Even with a strong drafter, a fixed speculation length $K$ can be suboptimal: large $K$ wastes computation when acceptance is low, while small $K$ caps potential speedup.
SpecDec++ \citep{specdec++} formulates the candidate length selection as an MDP, adaptively stopping drafting using an acceptance-prediction signal.
DISCO \citep{disco} similarly predicts when to stop speculation by dynamically selecting the speculation length.
PEARL \citep{pearl} addresses the system-level bottleneck of mutual waiting between drafting and verification by overlapping phases and enabling segmented, adaptive draft lengths.
Additionally, Block Verification \citep{block_verification} has been proposed to accelerate the verification phase itself.

\emph{Addressing training--inference misalignment.}
When a specialized drafter is trained, mismatches between training distributions and inference-time contexts can hurt acceptance.
HASS \citep{hass} proposes harmonized objectives and context alignment to reduce these inconsistencies.
CORAL \citep{coral} further improves consistency via cross-step representation alignment and reduces the effective cost of the LM head.
Orthogonally, Judge Decoding \citep{judge} observes that many rejected tokens are plausible and proposes relaxing verification via a compact judging module.

\paragraph{Tree-based speculative decoding.}
Tree-based Speculative Decoding (SD) generalizes the draft chain into a \emph{draft tree}, enabling the verifier to choose among multiple candidate branches. This significantly reduces the penalty of early mismatches. The key enabler is \emph{tree attention}, which allows the target model to verify multiple paths in parallel \citep{specinfer, medusa}.

\emph{Token-tree verification and structured drafting.}
SpecInfer \citep{specinfer} is a representative early system that organizes drafter outputs into a token tree verified via tree-based attention.
EAGLE \citep{eagle} advances this by performing autoregression at the feature level.
EAGLE-2 \citep{eagle2} introduces a context-aware dynamic draft tree guided by drafter confidence, while EAGLE-3 \citep{eagle3} further improves drafting via multi-layer fusion and training-time techniques.

\emph{Training-free tree construction via retrieval.}
To avoid training specialized drafters, some methods combine SD with retrieval.
REST \citep{rest} retrieves draft tokens from a datastore, while Prompt Lookup Decoding \citep{pld} leverages n-gram matching within the current context.
Token Recycling \citep{tokenrecycling} stores previously observed candidate-token transitions in a compact adjacency matrix to retrieve a draft tree.
SAM-Decoding \citep{samdecoding} utilizes a suffix automaton to efficiently find the longest suffix match for drafting. LogitSpec \citep{logitspec} proposes to use the last logit as a guidance to retrieve more matched and accurate draft tokens.
While plug-and-play, these methods are bounded by the availability of high-quality matches in the context or datastore.
Alternative draft-model-free approaches like Lookahead Decoding \citep{lade} generate parallel n-grams using Jacobi iteration, and EESD \citep{eesd} uses early exiting from intermediate layers to generate drafts.

\emph{Tree-attention efficiency.}
Since tree-based SD relies on non-trivial KV updates, system efficiency is critical. DeFT \citep{flash-tree-attn} proposes an IO-aware flash tree-attention algorithm tailored for tree-structured inference to improve verification throughput.

\paragraph{Summary and positioning of \method.}\label{talon_position}
While tree-based SD has become a standard paradigm, existing approaches face distinct limitations in adaptability and efficiency.
Learning-based methods, such as C2T \citep{c2t} and AdaEAGLE \citep{adaeagle}, require training specialized modules and remain constrained by rigid parameter spaces (e.g., predicting a fixed $K$), failing to achieve fully elastic topology changes.
Optimization-based methods also exhibit drawbacks: Sequoia \citep{sequoia} relies on \emph{offline} algorithms to find a globally optimal tree, resulting in a \emph{static} template that cannot adapt to instance-wise difficulty.
Similarly, search-based strategies like OPT-Tree \citep{opt-tree} often follow a ``generate-then-prune'' paradigm or employ complex search heuristics at inference time, which introduces non-trivial computational overhead.
\method distinguishes itself through a \emph{training-free, budget-driven} framework that operates via \emph{on-the-fly} construction.
Analogous to ``pre-pruning'' in decision trees, \method adopts a ``prune-while-expanding'' strategy: it iteratively allocates the node budget based on real-time confidence, naturally halting expansion in uncertain branches without generating wasteful nodes first.
This allows the draft tree to fluidly morph between ``deep-and-narrow'' and ``shallow-and-wide'' shapes, maximizing speculation utility with minimal construction cost.

\section{LLM Usage}
We used a large language model (LLM)–based writing assistant solely for grammar and wording improvements on draft text. The LLM did not generate research ideas, claims, proofs, algorithms, code, figures, or analyses, and it did not have access to any non-public data. All edits suggested by the LLM were manually reviewed and either accepted or rewritten by the authors, who take full responsibility for the final content. The LLM is not an author of this paper.
\end{document}